\documentclass[letterpaper]{article}
\usepackage[preprint]{aaai2027}
\usepackage[hyphens]{url}
\usepackage{graphicx}
\usepackage{natbib}
\usepackage{caption}
\usepackage{amsmath}
\usepackage{amsfonts}
\usepackage{amssymb}
\usepackage{booktabs}
\usepackage[table]{xcolor}
\usepackage{xurl}
\title{DemoEvolve: Demonstration-Guided Harness Evolution under Sparse Feedback}
\author{%
Lirong Che\textsuperscript{1,2} \quad
Yuzhe Yang\textsuperscript{2} \quad
Peiwen Lin\textsuperscript{2} \quad
Xu Cao\textsuperscript{2}\\
Chuang Wang\textsuperscript{2} \quad
Xueqian Wang\textsuperscript{2} \quad
Jian Su\textsuperscript{2}
}
\affiliations{%
\textsuperscript{1}Tsinghua University \quad
\textsuperscript{2}AgiBot
}

\begin{document}

\maketitle
\pagestyle{plain}
\thispagestyle{plain}

\begin{abstract}
Harness evolution enables frozen language model agents to adapt to unfamiliar tasks by modifying the
external programs that govern their behavior. For long-horizon tasks, each rollout is costly, and a
limited interaction budget may yield few examples of effective behavior. Sparse, delayed feedback
also makes it difficult for a coding agent to diagnose failures and determine which modifications
will improve performance. We present DemoEvolve, which uses human demonstrations to guide harness
evolution under limited interaction budgets. The coding agent examines demonstrations alongside its
own rollout history, extracts strategies and the conditions under which they apply, and turns them
into reusable harness components. We compare demonstration guidance with self-rollout evolution and
augmentation with retrieved textual knowledge under the same evolution procedure and agent
interaction budget. On held-out seeds, DemoEvolve improves mean capped final-round progress from
16.83 to 20.00 in Balatro and mean floor reached from 18.17 to 28.83 in Slay the Spire 2, compared
with the respective base harnesses. These results suggest that demonstrations can help harness
evolution make better use of limited interaction experience by providing concrete guidance for
diagnosis and modification.
\end{abstract}

\section{Introduction}
Large language model (LLM) agents can execute increasingly complex
tasks~\citep{yao2022react,wangVoyagerOpenEndedEmbodied2023a,jimenezSWEbenchCanLanguage2023},
yet most still cannot systematically learn from experience acquired after
deployment~\citep{heEvoTest2025,kartenContinualHarness2026}.
Humans, by contrast, acquire unfamiliar skills through repeated attempts, feedback, and
strategy revision. Harness evolution offers a promising mechanism for this form of test-time
learning: rather than updating model weights, it optimizes the external program around a
frozen model~\citep{lou2026autoharnessimprovingllmagents,leeMetaHarnessEndtoEndOptimization2026,linAgenticHarnessEngineering2026}.
Recent work has used this broader paradigm of optimizing external agent programs to
advance the performance frontier within established task domains, including text reasoning
and agentic coding~\citep{huAutomatedDesignAgentic2024,zhangAFlowAutomatingAgentic2024,leeMetaHarnessEndtoEndOptimization2026,linAgenticHarnessEngineering2026,xia2025livesweagent}.
We investigate an equally compelling direction: can harness evolution expand the range of
tasks an agent can master after deployment, enabling a general-purpose agent to acquire new
competence through limited practice and consolidate the resulting strategies into a reusable
harness?

\begin{figure}[t]
\centering
\includegraphics[width=\columnwidth]{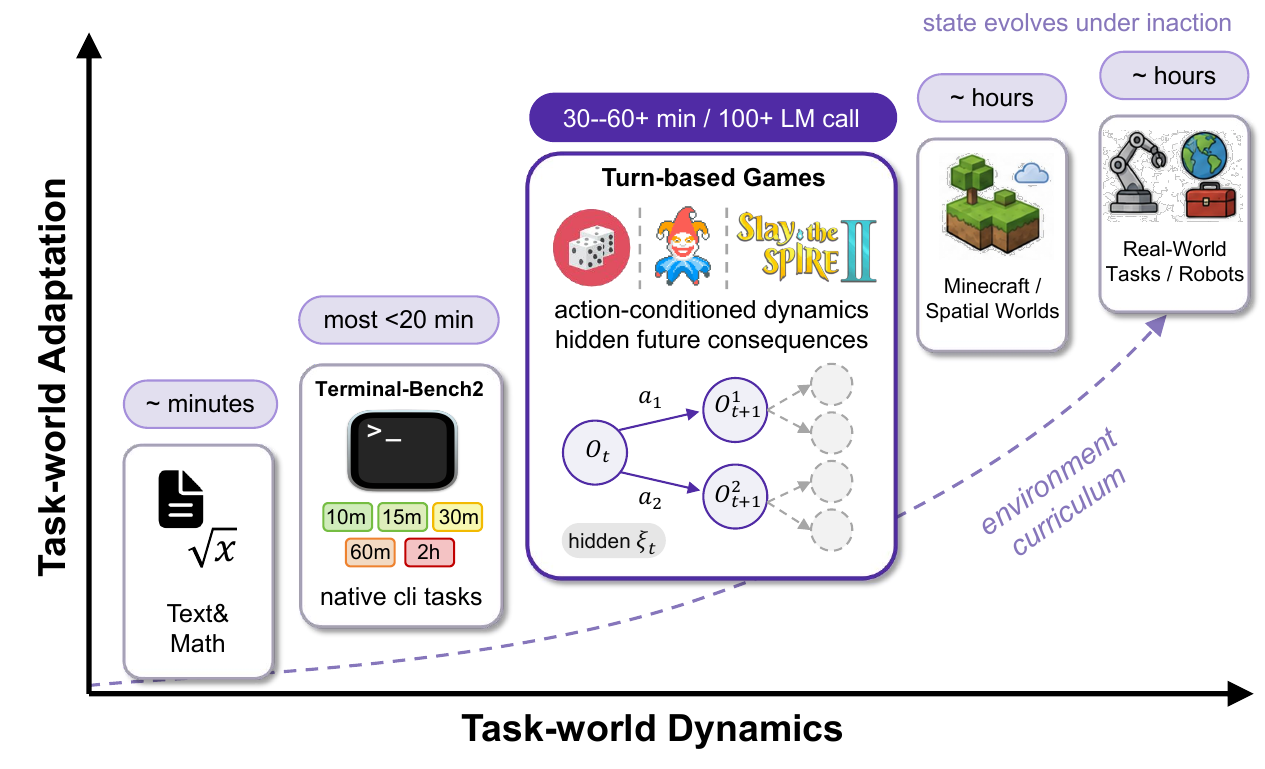}
\caption{\textbf{Games as a controlled testbed for task-world adaptation.}
Turn-based games combine action-conditioned, seed-dependent dynamics and sparse
long-horizon feedback with reproducibility through fixed seeds and explicit action
interfaces.}
\label{fig:task-regime}
\end{figure}

Games provide a controlled setting for studying this form of post-deployment
learning~\citep{guertler2025textarena,paglieri2024balrog,li2025textatari,hu2024gamearena,park2025orak}.
Unlike static text or reasoning tasks, they require agents to act in an evolving world: early
decisions shape future states, rewards arrive with delay, and successful behavior depends on
task-specific state abstractions and strategies. At the same time, fixed seeds and explicit
action interfaces make repeated evaluation reproducible and systematic. Turn-based games
therefore occupy a useful middle ground between model-native tasks and open-ended
real-world environments: they support long-horizon interaction while keeping iterative
harness evolution tractable~\citep{heEvoTest2025,kartenContinualHarness2026}.
Moreover, the same game-side interfaces used for agent interaction can record human
observations and actions during normal play, making aligned trajectories inexpensive to
collect.
Figure~\ref{fig:task-regime} illustrates this role.

This setting raises a natural question: how can harness evolution efficiently transform
limited interaction experience into reusable task competence? Existing methods operate
largely within a closed loop of self-generated rollouts, requiring the proposer to infer
improvement directions from the agent's own
attempts~\citep{leeMetaHarnessEndtoEndOptimization2026,linAgenticHarnessEngineering2026,heEvoTest2025,kartenContinualHarness2026}.
In long-horizon stochastic worlds, however, each expensive self-rollout often terminates in a
sparse, high-variance outcome that obscures which capability is missing and where the
harness should intervene. Humans learn unfamiliar tasks not only through trial and error,
but also by consulting external knowledge and observing competent behavior. We therefore
study whether expanding the experience available to the proposer can make harness evolution
more effective under a limited interaction budget.

We study demonstration-guided harness evolution, which we call \textbf{DemoEvolve},
by comparing three experience sources under the same automated evolution loop:
self-generated rollouts, external textual knowledge, and trajectory-level human
demonstrations. Unlike textual knowledge, which may describe a strategy without revealing
when it applies, demonstrations preserve the alignment among states, actions, and
outcomes~\citep{ross2011dagger,hester2018deep,nair2018overcoming}.
In the demonstration-guided condition, the proposer uses this aligned experience as
diagnostic evidence,
allowing proficient strategies to be consolidated into reusable, state-conditioned harness
mechanisms. Experiments on Liar's Dice~\citep{guertler2025textarena},
Balatro~\citep{coder2026balatrollm,coder2026balatrobench}, and Slay the
Spire~2~\citep{megacrit2026slaythespire2} show that self-rollout evolution is effective when
feedback is attributable but becomes unreliable under sparse, long-horizon feedback;
demonstration-guided evolution instead produces more effective and auditable harness
modifications.

Our contributions are summarized as follows:
\begin{itemize}
    \item We use interactive games to study harness evolution and identify the limitations
    of self-rollout feedback in long-horizon tasks.
    \item We study how trajectory-level demonstrations can guide automated harness
    evolution in \textbf{DemoEvolve}.
    \item We show that demonstrations enable more effective and auditable harness
    modifications than self-rollouts or textual knowledge.
\end{itemize}
\section{Related Work}
\label{sec:related-work}

\subsection{Agent Harness Evolution}

Beyond model weights, LLM-agent performance depends on task-specific \emph{harnesses}
governing prompts, workflows, tools, memory, and execution logic.
Recent work treats such structures as searchable or editable objects, optimizing prompts, workflows, tools, memories, or full agent harnesses through feedback-driven rewriting, program search, candidate selection, or outer-loop evolution~\citep{huAutomatedDesignAgentic2024,zhangAFlowAutomatingAgentic2024,openevolve,novikovAlphaEvolveCodingAgent2025a,lou2026autoharnessimprovingllmagents,leeMetaHarnessEndtoEndOptimization2026,linAgenticHarnessEngineering2026,robeyns2025selfimprovingcodingagent,zhang2026darwingodelmachineopenended,wang2025huxleygodelmachine,xia2025livesweagent}.
A related line of work improves agents by converting interaction histories into reusable reflections, memories, or skills that can be retrieved in later tasks~\citep{shinnReflexionLanguageAgents2023c,madaanSelfRefineIterativeRefinement2023c,zhaoExpeLLLMAgents2024,wangVoyagerOpenEndedEmbodied2023a,ye2026meta,zhangAgenticContextEngineering2025a,zhang2025memevolve,Xiong2026LearningTC,maSkillClawLetSkills2026b,xiaSkillRLEvolvingAgents2026c}.
Most closely related, EvoTest formulates test-time learning as evolutionary optimization of
an agent's full configuration between consecutive episodes, using prior interaction
trajectories and outcomes as feedback~\citep{heEvoTest2025}.
These works assume that past experience contains reusable improvement signals; we instead
ask when self-generated experience remains useful in non-model-native, long-horizon tasks
with sparse, noisy, and hard-to-attribute failures.

\subsection{Experience Shaping}

The form and distribution of experience can strongly affect learning and adaptation.
Imitation learning and learning from demonstrations use expert trajectories to reduce the
burden of unguided exploration and mitigate distribution mismatch between expert states and
learner-induced states~\citep{ross2011dagger,ho2016generative,hester2018deep,nair2018overcoming,rajeswaran2018learning}.
Curriculum and self-paced learning further show that controlling the order, difficulty, or
goals of experience can improve optimization stability and sample efficiency
~\citep{bengio2009curriculum,kumar2010selfpaced,florensa2017reverse}.
Exploration methods for sparse-reward tasks similarly emphasize reaching informative or
promising states rather than exploring uniformly from scratch~\citep{ecoffet2021first}.
These lines of work inspire our investigation of how different experience sources can guide
harness evolution.

\subsection{LLM Agents in Games}

Games provide a complementary testbed to static question answering and offline benchmarks.
Recent benchmarks place language agents in text games, video games, board and card games, and multi-agent environments to evaluate dynamic state tracking, long-horizon planning, hidden-information reasoning, social reasoning, and opponent modeling~\citep{guertler2025textarena,paglieri2024balrog,li2025textatari,hu2024gamearena,duan2024gtbench,park2025orak,fu2026catarenaevaluatingevolutionarycapabilities}.
Related work has also begun to use stochastic roguelike deck-building games to expose limitations of LLM agents in long-term resource management, combinatorial strategy, and risk control~\citep{coder2026balatrollm,coder2026balatrobench}.
Unlike work that primarily evaluates fixed agents or trains game-playing policies, we use games as test environments for harness self-evolution under stochasticity, long-horizon decisions, and dynamic interaction.
\section{Method}
\label{sec:method}

\subsection{Preliminaries}
\label{sec:method-problem}

We formulate harness evolution as an adaptation mechanism for frozen language-model agents
in unfamiliar task worlds. Let \(W_T\) denote a task world, \(x\sim\mathcal{X}_T\) an
episode instance, \(\pi_\phi\) a frozen model, and \(h\in\mathcal{H}_T\) a task-oriented
harness. The task world exposes observations and task-specific action interfaces, while
the harness determines how information is processed and presented to the model, including
state abstraction, memory, tool usage, action interfaces, and execution logic. In games,
these actions may include operations such as playing, discarding, buying, selling, or
rerolling.

Together, \(W_T\), \(\pi_\phi\), and \(h\) induce a rollout distribution over interaction
trajectories. The objective of harness evolution is to discover a task-specific harness that
maximizes performance without updating the underlying model:
\[
h^*
=
\arg\max_{h\in\mathcal{H}_T}
\mathbb{E}_{x\sim\mathcal{X}_T,\,
\tau\sim p_{W_T,\pi_\phi,h}(\cdot\mid x)}
\left[
R_T(\tau,x)
\right].
\]
Here, \(\tau=(o_0,a_0,o_1,a_1,\ldots,o_T)\) denotes the task-world interaction trajectory
and \(R_T(\tau,x)\) is the task reward, which is often sparse and observed only after the
rollout. The model parameters \(\phi\) remain fixed throughout; all adaptation is achieved
through modifications to the executable harness \(h\).

\paragraph{Controlled task-world dynamics.}
Our goal is to study harness evolution in task worlds that are dynamic enough to require
state abstraction and long-horizon reasoning, while remaining controlled enough for
systematic search and evaluation. We therefore distinguish the evaluator's view of a
fixed-seed simulator from the agent-facing view exposed through the harness. Let
\(\tilde{s}_t\) denote the complete task-world state, let
\(o_t=\Omega(\tilde{s}_t)\) be the observation available to the agent, and let \(\rho\)
denote an episode seed. Under a fixed seed, the full-state transition may be deterministic:
\[
\tilde{s}_{t+1}=F_\rho(\tilde{s}_t,a_t),
\]
allowing repeated evaluation on the same seed to be reproducible.

Reproducibility for the evaluator, however, does not imply that the agent knows the future
consequences of its actions. The agent acts based on its observation-action history
\(\eta_t=(o_0,a_0,\ldots,o_t)\), rather than the complete task-world state. Therefore, the
next observation remains uncertain from the agent's perspective:
\[
\begin{aligned}
p_\rho(o_{t+1}\mid \eta_t,a_t)
&=
\sum_{\tilde{s}\in\mathcal{S}}
p_\rho(\tilde{s}_t=\tilde{s}\mid \eta_t)
\\
&\quad \cdot
\mathbf{1}\!\left[
o_{t+1}=\Omega(F_\rho(\tilde{s},a_t))
\right].
\end{aligned}
\]

We refer to this uncertainty induced at the observation and decision interface as
\emph{agent-facing uncertainty}, rather than intrinsic nondeterminism after all simulator
variables are fixed. Turn-based games therefore provide a controlled regime for studying
task-world adaptation: they introduce action-conditioned future consequences, hidden
state, and delayed feedback, while retaining fixed seeds and explicit action interfaces for
reproducible harness evolution.

\paragraph{Diagnosing harness interventions.}
The resulting optimization problem differs from standard game learning. In reinforcement
learning, sparse outcomes are typically connected to policy or value updates through a
predefined learning rule. In our setting, the optimized object is a non-differentiable
executable harness whose components may include prompts, state abstractions, memory
mechanisms, tool interfaces, and control flow. A terminal reward therefore does not provide
a direct update target for a particular harness modification.

Instead, the proposer must infer which capabilities are missing and where harness
interventions should be introduced based on trajectories, execution logs, and sparse
outcomes. This diagnosis becomes challenging when feedback is noisy and delayed. A failed
run may indicate insufficient state abstraction, poor action prioritization, weak resource
management, inappropriate tool usage, or ineffective control logic. Moreover, a candidate
harness may appear better because of trajectory variance or seed-specific effects rather
than because its modification provides a genuine capability improvement. We therefore
distinguish \emph{score improvements} from \emph{functional harness improvements}: the
latter require the modified harness to produce auditable changes in the model-facing
interface or execution behavior.

\begin{figure*}[!t]
    \centering
    \includegraphics[width=0.7\textwidth]{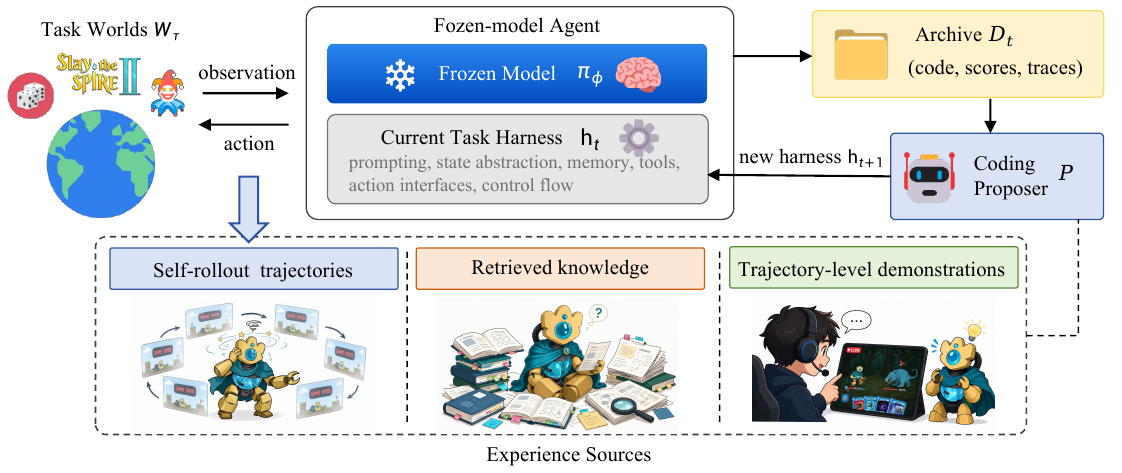}
    \caption{
    \textbf{Experience sources for task-world adaptation.}
    A coding proposer \(P\) evolves task-specific harness \(h_t\) from prior harnesses,
    scores, and traces, with access to self-generated trajectories, retrieved knowledge, or
    human demonstrations.}
    \label{fig:overview}
\end{figure*}

\subsection{Experience Sources for Harness Evolution}
\label{sec:method-information-regimes}

We study how different forms of experience influence harness evolution for task-world
adaptation using an outer-loop coding proposer \(P\) that iteratively improves
a task-specific harness around a frozen model. The proposer is responsible for diagnosing
failure modes, identifying missing capabilities, and generating executable harness
modifications; it is not the task-execution agent itself.

Figure~\ref{fig:overview} summarizes the outer-loop process and the three experience sources.

We intentionally adopt a minimal evolution procedure to isolate the effect of experience
sources. The proposer itself determines which previous candidates, trajectories, and
execution artifacts are relevant for diagnosis and modification. We do not introduce
additional search heuristics or manually designed candidate-selection rules. Thus, all
experimental conditions share the same proposer, evolution loop, search budget, and
evaluation protocol, differing only in the additional experience available to the proposer.

At each iteration, the proposer accesses an evolution history
\(\mathcal{D}_t\) containing previously evaluated harnesses and their associated artifacts.
Each evaluated harness contributes its source code, development-set scores, rollout
trajectories, and raw execution traces. Rollout trajectories capture environment-level
interactions, while execution traces record harness-level behaviors, including model
inputs, model outputs, parsed actions, tool calls, errors, and intermediate states. The
proposer queries \(\mathcal{D}_t\) through standard file operations rather than receiving a
manually constructed summary, allowing it to selectively inspect relevant evidence during
harness diagnosis.

For outer-loop iteration \(t\), let \(k_t\) denote the number of candidate harnesses proposed
by the proposer. The evolution process is formulated as:
\[
\begin{aligned}
\{h^{(t+1,j)}\}_{j=1}^{k_t}
&\leftarrow P(I_t),\\
\mathcal{D}_{t+1}
&=
\mathcal{D}_{t}
\cup
\{\mathrm{EvalDir}(h^{(t+1,j)})\}_{j=1}^{k_t},
\end{aligned}
\]
where \(I_t\) denotes the information available to the proposer. Each candidate is evaluated
on development instances, and the resulting artifacts are added to the evolution history.
Final evaluation instances remain hidden throughout evolution.

Humans acquire competence in unfamiliar environments through different sources of
experience: they may practice through trial and error, consult external knowledge, or
observe expert behavior. Inspired by this observation and by learning from demonstrations
in sequential decision-making, we investigate three experience sources for harness
evolution:
\[
\begin{aligned}
I_t^{\mathrm{self}}
&=
\mathcal{D}_t,\\
I_t^{\mathrm{know}}
&=
\mathcal{D}_t \cup K^{\mathrm{text}},\\
I_t^{\mathrm{demo}}
&=
\mathcal{D}_t \cup T^{\mathrm{human}}.
\end{aligned}
\]

All three settings use the same automated evolution procedure. They differ only in the
additional experience signal provided to the proposer.

\paragraph{Self-generated experience.}
In this setting, the proposer relies only on its own evolution history
\(\mathcal{D}_t\), including previous harness implementations, rollout trajectories,
evaluation outcomes, and execution traces. This corresponds to autonomous trial-and-error
adaptation, where the agent improves solely from its own interaction experience.

\paragraph{Retrieved knowledge.}
In this setting, the proposer additionally has access to external textual knowledge
\(K^{\mathrm{text}}\). Rather than providing a manually curated knowledge base, we allow the
coding proposer to actively retrieve relevant information when diagnosing failures. The
retrieved information may include task rules, tutorials, strategy guides, or reference
materials. This setting evaluates whether general task knowledge can accelerate harness
adaptation when an agent encounters an unfamiliar task world.

\paragraph{Human demonstrations.}
In this setting, the proposer additionally receives human interaction trajectories:
\[
\begin{aligned}
T^{\mathrm{human}}
&=
\{(\tau_j^H,C_j^H)\}_{j=1}^{m},\\
\tau_j^H
&=
(o_0^H,a_0^H,o_1^H,a_1^H,\ldots,o_{T_j}^H),
\end{aligned}
\]
where \(\tau_j^H\) denotes a human gameplay trajectory and
\(C_j^H\) contains optional human-provided commentary recorded during interaction.

\begin{figure*}[!t]
    \centering
    \includegraphics[width=0.8\textwidth]{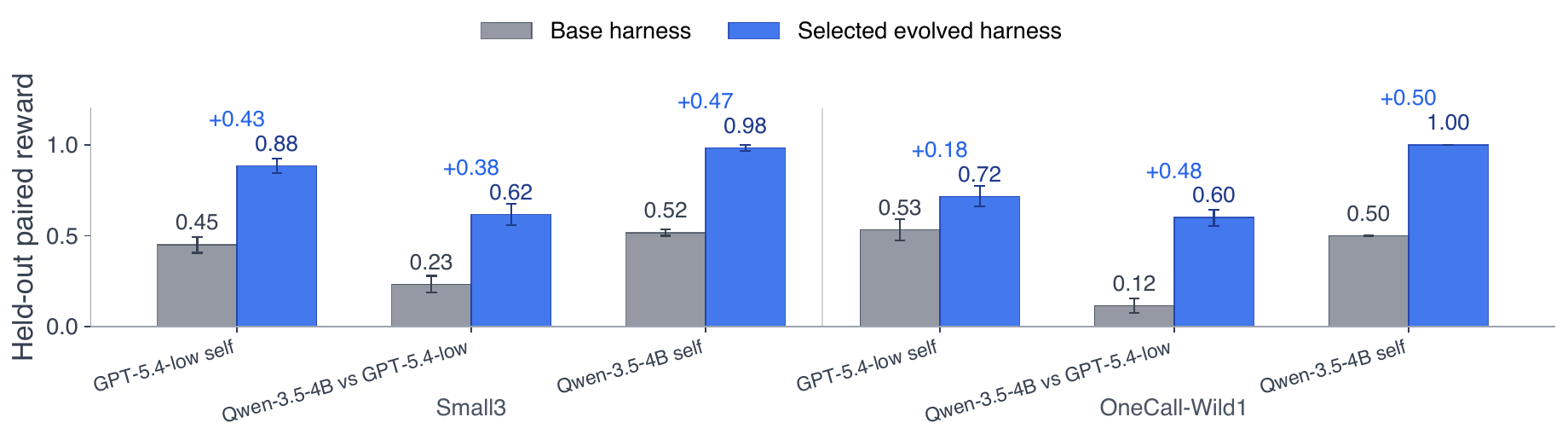}
    \caption{Held-out TextArena reward after development-only harness selection on 30 paired
    logical seeds (mean \(\pm\) standard error). Numbers above pairs show
    evolved-minus-base differences.}
    \label{fig:textarena-heldout-generalization}
\end{figure*}

Unlike action-level supervision in imitation learning, human commentary is not aligned with
individual actions. Instead, it provides trajectory-level context about strategies,
intentions, observations, and decisions throughout the interaction. The coding proposer
uses these demonstrations as diagnostic evidence by comparing competent human behavior
with failed agent trajectories. Based on this evidence, it autonomously determines where
and how to introduce harness interventions, such as state abstractions, phase-specific
logic, resource-management strategies, or auxiliary computation tools.

\paragraph{Human trajectory acquisition.}
We instrument the original game interfaces to expose the same observation--action protocol
used by agents. Normal gameplay logs observations, actions, timestamps, and outcomes, with
optional text or voice commentary. This yields aligned trajectories without manual harness
design or fine-grained annotation; all harness modifications remain proposer-generated.
\section{Experiments}
\label{sec:experiments}

We organize our experiments around three research questions: \textbf{RQ1:} When is
self-rollout evolution sufficient? \textbf{RQ2:} Does demonstration guidance improve
long-horizon adaptation under limited interaction budgets? \textbf{RQ3:} What drives the
resulting gains from demonstration guidance?
% We first use
% Liar's Dice as a lightweight setting in which failures are sufficiently attributable for
% self-generated experience to guide harness evolution. We then evaluate DemoEvolve in
% Balatro, where long horizons and sparse outcomes make harness-level credit assignment more
% difficult. Finally, we examine whether the resulting improvements reflect the transfer of a
% strategy-level behavioral pattern.

\begin{table*}[!t]
\centering
\footnotesize
\renewcommand{\arraystretch}{1.08}
\begin{minipage}[t]{0.485\textwidth}
\centering
\setlength{\tabcolsep}{3.2pt}
\begin{tabular*}{\linewidth}{@{\extracolsep{\fill}}lccc@{}}
\toprule
\multicolumn{4}{c}{\textbf{BalatroBench}} \\
\cmidrule(lr){1-4}
Method & Dev. & Held-out & Overall \\
\midrule
\rowcolor{black!8}
\multicolumn{4}{l}{\textbf{\emph{Non-evolution Baseline}}} \\
BalatroLLM & 17.00 (33\%) & 16.83 (33\%) & 16.93 (33\%) \\
\midrule
\rowcolor{black!8}
\multicolumn{4}{l}{\textbf{\emph{Harness Evolution Methods}}} \\
Meta-Harness & 19.33 (44\%) & 16.33 (33\%) & 18.13 (40\%) \\
Meta-Harness-Web & 16.67 (22\%) & 17.00 (33\%) & 16.80 (27\%) \\
\textbf{DemoEvolve} & \textbf{23.33 (89\%)} & \textbf{20.00 (67\%)} & \textbf{22.00 (80\%)} \\
\bottomrule
\end{tabular*}
\end{minipage}
\hfill
\begin{minipage}[t]{0.485\textwidth}
\centering
\setlength{\tabcolsep}{3.2pt}
\begin{tabular*}{\linewidth}{@{\extracolsep{\fill}}lccc@{}}
\toprule
\multicolumn{4}{c}{\textbf{STS2Bench}} \\
\cmidrule(lr){1-4}
Method & Dev. & Held-out & Overall \\
\midrule
\rowcolor{black!8}
\multicolumn{4}{l}{\textbf{\emph{Non-evolution Baseline}}} \\
STS2LLM & 20.00 (33\%) & 18.17 (17\%) & 19.27 (27\%) \\
\midrule
\rowcolor{black!8}
\multicolumn{4}{l}{\textbf{\emph{Harness Evolution Methods}}} \\
Meta-Harness & 23.67 (44\%) & 23.17 (50\%) & 23.47 (47\%) \\
Meta-Harness-Web & 27.44 (67\%) & 24.83 (50\%) & 26.40 (60\%) \\
\textbf{DemoEvolve} & \textbf{30.33 (89\%)} & \textbf{28.83 (100\%)} & \textbf{29.73 (93\%)} \\
\bottomrule
\end{tabular*}
\end{minipage}
\caption{Long-horizon task-world adaptation. Entries report mean progress, with success
rates in parentheses rounded to the nearest percentage point, on development, held-out, and
all seeds. BalatroBench uses capped final-round progress, with game completion as success.
STS2Bench uses the game floor reached, with clearing Act~1 as success. Metrics are
comparable only within each panel.}
\label{tab:long-horizon-results}
\end{table*}

\subsection{RQ1: When Is Self-Rollout Evolution Sufficient?}
\label{sec:textarena-analysis}
\label{sec:textarena-rq1}

\paragraph{Setup.}
We evaluate self-rollout harness evolution in TextArena Liar's
Dice~\citep{guertler2025textarena}, a lightweight interactive game with stochastic rolls,
hidden information, opponent interaction, and sequential decisions. This environment serves
as a positive control: when rollout feedback is sufficiently attributable, the proposer should
be able to convert observed failures into executable harness structure around a frozen model.

We use two variants. Small3 is a three-dice sequential bidding game, while OneCall-Wild1 is
a five-dice single-call variant in which ones are wild. For each task, we evaluate three
target/opponent cells: GPT-5.4-low self-play, Qwen-3.5-4B against GPT-5.4-low, and
Qwen-3.5-4B self-play. Candidate harnesses are selected only from development rollouts.
Held-out evaluation uses disjoint seeds and paired-side control with 30 logical seeds per
comparison.

\paragraph{Results.}
The search loop finds higher-development-reward harnesses within a small number of
proposals across both tasks and all model pairings; full search curves are provided in the
appendix. As shown in Figure~\ref{fig:textarena-heldout-generalization}, the selected
harnesses improve held-out performance in all six comparisons. The task-balanced macro
reward increases from 0.392 to 0.800, showing that self-generated rollouts can provide a
useful evolution signal when failures are locally attributable.

\paragraph{Mechanism.}
The gains are not merely produced by longer prompts. The evolved harnesses add
task-specific structures such as observation parsing, legal-action guards, bid repair,
probability estimates, expected-value call thresholds, opening policies, and conservative
safety overrides. Qualitative cases in the appendix further show that evolution adapts the
model--harness boundary to the base model and task variant: weaker-model runs may become
near-symbolic threshold policies, while stronger-model runs retain a neuro-symbolic division
of labor in which code handles legality and probability and the model handles residual
strategic judgment.

\paragraph{Takeaway.}
Self-rollout evolution is sufficient when useful behavior can be expressed as compact local
heuristics and rollout traces expose enough information to diagnose failures. We next test
the limits of this pattern in long-horizon task worlds, where the value of an intervention is
more state-dependent and terminal outcomes provide a noisier search signal.

\subsection{RQ2: Does Demonstration Guidance Improve Long-Horizon Adaptation?}
\label{sec:long-horizon-results}
\label{sec:balatro-setup}

\paragraph{Setup.}
We evaluate demonstration-guided harness evolution in two long-horizon, turn-based
deck-building games: Balatro and Slay the Spire~2 (STS2). Balatro is a poker-based
deck-building roguelite~\citep{coder2026balatrollm,coder2026balatrobench}. Successful play
requires delayed resource management under seed-dependent card draws, shop inventories,
boss blinds, and joker synergies, making terminal feedback sparse, high-variance, and
difficult to attribute to individual harness mechanisms. STS2~\citep{megacrit2026slaythespire2}
additionally requires route planning, card acquisition, deck construction, combat decisions,
and resource management across a sequence of encounters. We include STS2 as a second,
independently instantiated task world to test whether the benefits of DemoEvolve extend
beyond Balatro-specific mechanics.

For Balatro, we follow the fixed-seed protocol of
BalatroBench~\citep{coder2026balatrobench}. As STS2 has no established agent benchmark,
we construct a matched protocol with the same number of seeds and the same
development--held-out split. For each game, 60\% of seeds are used during evolution, while
the remaining 40\% are reserved exclusively for final evaluation. Candidate harnesses are
evaluated and selected only on the development seeds, after which the selected harness is
re-evaluated from scratch on the full seed set. We match the outer-loop budget and evaluation
protocol across the two games. Full experimental details are provided in the appendix.

\paragraph{Compared conditions.}
We compare four conditions in each game. \textbf{Base Harness} uses the corresponding
fixed task-specific harness without outer-loop evolution. Both agents interact with their
game through game-side mods that expose structured observations and executable actions.
For Balatro, we use the publicly available \textbf{BalatroLLM}
harness~\citep{coder2026balatrollm}. For STS2, we build \textbf{STS2LLM (ours)} on an
adapted mod-based interface, following the overall agent architecture of
BalatroLLM. The corresponding task-specific harness serves as the common initialization
for all evolution conditions.

\textbf{Meta-Harness} uses Meta-Harness~\citep{leeMetaHarnessEndtoEndOptimization2026}
as the shared evolution backbone. Its coding proposer may inspect prior harness code,
development scores, self-generated rollouts, and raw execution traces before proposing new
candidates. Across both games, the task-execution model is GPT-5.4 with low reasoning
effort, while the coding proposer is Claude Opus 4.7 with maximum reasoning effort. Each
evolution condition runs for 10 outer-loop iterations.

\textbf{Meta-Harness-Web} is our knowledge-augmented variant of
Meta-Harness~\citep{leeMetaHarnessEndtoEndOptimization2026}. It retains the same
outer-loop evolution backbone and additionally equips the coding proposer with a web-search
tool. The evolution skill encourages the proposer to retrieve external task information,
including game rules, tutorials, strategy guides, and reference implementations.

\textbf{DemoEvolve} uses the same backbone and additionally provides human trajectories
represented in the same observation--action format as agent rollouts. The trajectories are
collected only on the development seeds and are available only to the coding proposer during
evolution; they are not inserted into the task-execution agent's context during final
evaluation.

Across the three evolution conditions, the initial harness, search budget, task-execution
model, coding proposer, selection rule, and evaluation protocol are held fixed. They differ
only in the auxiliary evidence available to the proposer.

\paragraph{Results.}
Table~\ref{tab:long-horizon-results} shows that DemoEvolve achieves the strongest
performance in both task worlds. Seed-stratified exact permutation tests with Holm
correction across the two primary environment-level comparisons confirm significant
improvements over the Base Harness in both Balatro
($p_{\mathrm{Holm}}=.023$) and STS2 ($p_{\mathrm{Holm}}<.001$). In Balatro, Meta-Harness improves on
the development seeds but does not transfer to held-out seeds, while Meta-Harness-Web
remains close to the base harness. A functional audit, detailed in the appendix, further
reveals that the dynamic hook added by the selected Meta-Harness candidate never triggers
because of an implementation error, so its apparent development gain is better interpreted
as noisy candidate selection than as a verified harness improvement.

STS2 exhibits a smoother improvement across the evolution conditions: Meta-Harness and
Meta-Harness-Web both outperform the base harness, while DemoEvolve produces the largest
gains on both development and held-out seeds. Across the two independently instantiated
task worlds, demonstration-guided evolution therefore provides the most consistent
long-horizon adaptation under the shared interaction budget.

\begin{table*}[!t]
\centering
\footnotesize
\setlength{\tabcolsep}{4.2pt}
\renewcommand{\arraystretch}{1.04}
\begin{tabular}{lcccccc}
\toprule
& \multicolumn{2}{c}{\textbf{Mechanism}} & \multicolumn{2}{c}{\textbf{BalatroBench}} & \multicolumn{2}{c}{\textbf{STS2Bench}} \\
\cmidrule(lr){2-3} \cmidrule(lr){4-5} \cmidrule(lr){6-7}
Variant & Prior & Evolution & Dev. & Held-out & Dev. & Held-out \\
\midrule
Base Harness & $\times$ & $\times$ & 17.00 (33\%) & 16.83 (33\%) & 20.00 (33\%) & 18.17 (17\%) \\
\hspace{0.6em}+ Segment Summary & $\checkmark$ & $\times$ & 20.56 (67\%) & 16.83 (17\%) & 14.89 (0\%) & 14.67 (0\%) \\
\hspace{0.6em}+ Episode Summary & $\checkmark$ & $\times$ & 16.56 (22\%) & 16.17 (17\%) & 21.78 (56\%) & 18.17 (17\%) \\
\textbf{DemoEvolve} & $\checkmark$ & $\checkmark$ & \textbf{23.33 (89\%)} & \textbf{20.00 (67\%)} & \textbf{30.33 (89\%)} & \textbf{28.83 (100\%)} \\
\bottomrule
\end{tabular}
\caption{Demonstrations without evolution. Summary variants inject information from the
same nine demonstrations into the fixed base harness. Prior and Evolution denote access to
demonstration-derived context and outer-loop evolution, respectively. Metrics and success
definitions follow Table~\ref{tab:long-horizon-results}.}
\label{tab:demo-without-evolution}
\end{table*}

\begin{figure*}[!t]
    \centering
    \includegraphics[width=0.9\textwidth]{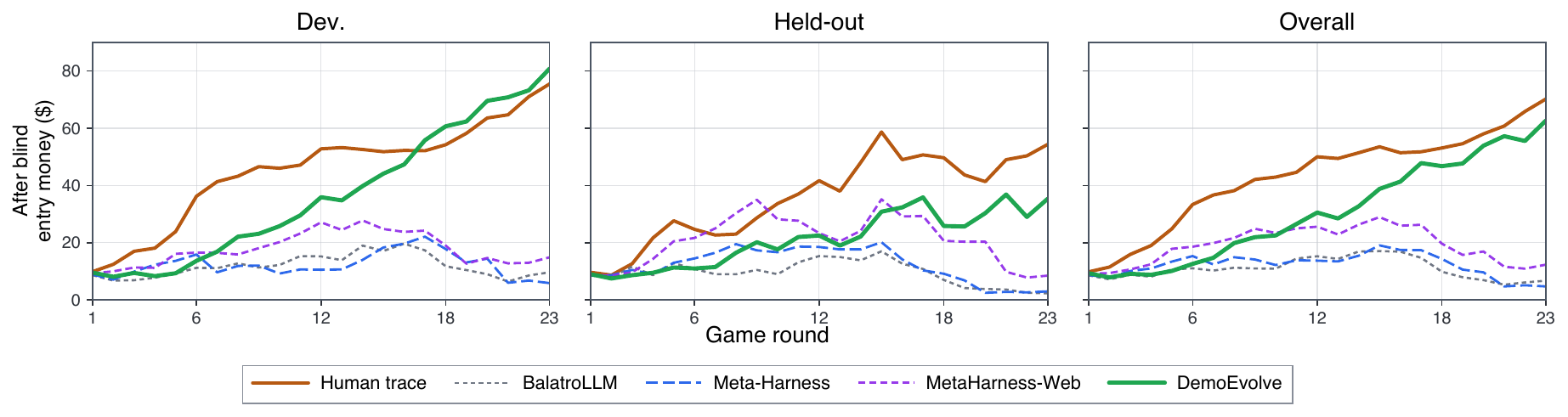}
    \caption{Balatro after-blind entry money on development, held-out, and all seeds. Human
    trajectories are shown only as a post-hoc reference; held-out demonstrations are never
    available during evolution.}
    \label{fig:balatro_attrition_adjusted_economy}
\end{figure*}

\paragraph{Key insight.}
These results suggest that experience sources differ not only in how much task information
they provide, but also in their diagnostic granularity. Self-generated rollouts offer
retrospective evidence of failure without showing a competent alternative, while retrieved
knowledge supplies general strategic principles but often omits the state boundaries under
which those principles should apply. Both sources may therefore inspire directionally
correct harness modifications whose triggering conditions are poorly localized.
Trajectory-level demonstrations preserve the alignment among states, actions, and outcomes,
allowing the proposer to infer not only \emph{what} should change, but also \emph{where} the
intervention belongs and \emph{when} it should activate. Section~\ref{sec:gain-analysis}
examines this distinction concretely through economy management in Balatro.

\subsection{RQ3: What Drives the Gains of DemoEvolve?}
\label{sec:gain-analysis}

\subsubsection{Disentangling Demonstrations from Evolution}
\label{sec:disentangling-demonstrations}

A natural interpretation of Table~\ref{tab:long-horizon-results} is confounded: relative to
the base harness, DemoEvolve differs both in the information it receives---expert
trajectories---and in the mechanism used to consume that information---the outer evolution
loop. Its gains could therefore reflect only the value of demonstrations as an additional
prior, rather than their value for harness evolution.

To isolate this mechanism, we introduce a \emph{demonstrations without evolution}
condition. The fixed baseline harness receives information derived from the same nine
demonstrations available to DemoEvolve, but the outer evolution loop is removed entirely.
Directly inserting the raw trajectories is infeasible: the nine demonstrations contain 3,080
decisions, approximately 68 times the context budget of a single model call. We therefore
evaluate two complementary injection mechanisms that bracket the design space.
\emph{Episode-level summaries} compress each trajectory independently, preserving the
coherent plan within an episode; all nine summaries are injected together. \emph{Segment-level
summaries} instead compress across the temporal dimension, preserving fine-grained guidance
for corresponding decision stages. We construct eight Balatro segments and six STS2 segments,
with construction details provided in the supplementary material, and conditionally inject only
the summary corresponding to the agent's current phase.

Table~\ref{tab:demo-without-evolution} shows that, under the two summary-injection schemes
considered here, direct demonstration conditioning does not reproduce DemoEvolve's held-out
gains. Both summary baselines match or underperform the base harness on held-out seeds,
whereas DemoEvolve remains strongest in both games. This suggests that its gains arise from
converting trajectory evidence into reusable harness mechanisms through evolution, rather
than from demonstration-derived advice alone.

\subsubsection{Strategy-Level Behavioral Changes}
\label{sec:balatro_behavior_evidence}

To identify strategy-level behavioral changes, we analyze economy management in Balatro.
Exact action matching is unreliable because action values depend strongly on the current
build and game state. We therefore track after-blind entry money as a coarse probe of how
agents balance saving against investment. Runs that terminate before a round contribute
zero, so late-round values jointly reflect survival and retained economy; full metric
definitions and numerical summaries are provided in the appendix.

Figure~\ref{fig:balatro_attrition_adjusted_economy} shows that Meta-Harness initially tracks
human-like economy levels but diverges in later rounds, producing little overall change
relative to BalatroLLM. Meta-Harness-Web moves closer to the human curve overall, but does
not reliably convert economy preservation into late survival. DemoEvolve reduces both
overall and late-game distance to the human economy curve while maintaining high late reach.
This supports a strategy-level transfer interpretation: demonstrations help the evolved
harness sustain a reusable resource-management strategy rather than merely improve final
scores through noisy candidate selection.

Economy management also illustrates the mechanism behind this result. Self-rollout
diagnoses can identify symptoms such as overspending, lost interest bonuses, or weak
late-round buffers, but often translate them into broad purchase-side safeguards. Such
safeguards may produce the early human-like economy seen with Meta-Harness, yet fail to
adapt as the value of spending changes later in a run. Retrieved knowledge can correctly
emphasize the importance of maintaining an economy, but static advice to save money may
make the agent overly conservative and prevent purchases needed for survival. Human
trajectories instead reveal the state-dependent boundary between saving and investing,
enabling the proposer to encode economy guidance as a selectively triggered harness
mechanism rather than a persistent behavioral bias.

\FloatBarrier
\section{Conclusion}
\label{sec:conclusion}

We studied harness evolution for frozen-model agents under sparse feedback. Our experiments
show that self-generated trajectories can support effective evolution when feedback is
readily attributable, but become vulnerable to sparse rewards and noisy candidate selection
in long-horizon stochastic tasks. In the demonstration-guided condition, human trajectories
preserve state--action correspondences, helping the coding proposer translate competent behavior
into reusable, state-conditioned harness mechanisms. Across two long-horizon task worlds,
this approach yields more consistent generalization to held-out instances.

\section{Limitations}
\label{sec:limitations}

Our study focuses on turn-based games, the controlled middle ground shown in
Figure~\ref{fig:task-regime}. More dynamic worlds may continue to evolve while the agent
reasons, use continuous or asynchronous interfaces, and permit less reproducible evaluation.
Our results are therefore a first step toward broader task-world adaptation, not evidence of
direct transfer to all interactive environments.

Our study also assumes that human demonstrations can be aligned with the agent's
observation--action interface. Instrumented game mods make such data inexpensive to collect,
but alignment may be difficult when humans and agents use different interfaces. Future work
should study cross-interface trajectory alignment and robustness to imperfect demonstrations.

\bibliography{references}

@misc{heEvoTest2025,
  title = {{EvoTest}: Evolutionary Test-Time Learning for Self-Improving Agentic Systems},
  author = {He, Yufei and Liu, Juncheng and Liu, Yue and Li, Yibo and Cao, Tri and Hu, Zhiyuan and Xu, Xinxing and Hooi, Bryan},
  year = {2025},
  eprint = {2510.13220},
  archivePrefix = {arXiv},
  primaryClass = {cs.AI}
}

@misc{kartenContinualHarness2026,
  title = {Continual Harness: Online Adaptation for Self-Improving Foundation Agents},
  author = {Karten, Seth and Zhang, Joel and Upaa, Tersoo and Feng, Ruirong and Li, Wenzhe and Shi, Chengshuai and Jin, Chi and Vodrahalli, Kiran},
  year = {2026},
  eprint = {2605.09998},
  archivePrefix = {arXiv},
  primaryClass = {cs.LG}
}

@misc{megacrit2026slaythespire2,
  title = {Slay the Spire 2},
  author = {{Mega Crit}},
  year = {2026},
  howpublished = {\url{https://www.megacrit.com/press-kits/slay-the-spire-2/}},
  note = {Accessed July 27, 2026}
}

@article{zhang2025memevolve,
    title   = {Memevolve: Meta-evolution of agent memory systems},
    author  = {Zhang, Guibin and Ren, Haotian and Zhan, Chong and Zhou, Zhenhong and Wang, Junhao and Zhu, He and Zhou, Wangchunshu and Yan, Shuicheng},
    journal = {arXiv preprint arXiv:2512.18746},
    year    = {2025}
}

@misc{Xiong2026LearningTC,
  title = {Learning to Continually Learn via Meta-learning Agentic Memory Designs},
  author = {Xiong, Yiming and Hu, Shengran and Clune, Jeff},
  year = {2026},
  eprint = {2602.07755},
  archivePrefix = {arXiv},
  primaryClass = {cs.AI},
  doi = {10.48550/arXiv.2602.07755},
  url = {https://arxiv.org/abs/2602.07755}
}

@misc{openevolve,
    title     = {OpenEvolve: an open-source evolutionary coding agent},
    author    = {Asankhaya Sharma},
    year      = {2025},
    howpublished = {\url{https://github.com/algorithmicsuperintelligence/openevolve}},
    note = {GitHub repository},
    url       = {https://github.com/algorithmicsuperintelligence/openevolve}
}

@inproceedings{jimenezSWEbenchCanLanguage2023,
  title = {{SWE-bench}: Can Language Models Resolve Real-World {GitHub} Issues?},
  shorttitle = {SWE-Bench},
  booktitle = {The Twelfth International Conference on Learning Representations},
  author = {Jimenez, Carlos E. and Yang, John and Wettig, Alexander and Yao, Shunyu and Pei, Kexin and Press, Ofir and Narasimhan, Karthik R.},
  year = 2024,
  url = {https://openreview.net/forum?id=VTF8yNQM66},
  urldate = {2026-04-27},
  langid = {english}
}

@inproceedings{madaanSelfRefineIterativeRefinement2023c,
  title = {Self-Refine: Iterative Refinement with Self-Feedback},
  shorttitle = {Self-Refine},
  booktitle = {Thirty-Seventh Conference on Neural Information Processing Systems},
  author = {Madaan, Aman and Tandon, Niket and Gupta, Prakhar and Hallinan, Skyler and Gao, Luyu and Wiegreffe, Sarah and Alon, Uri and Dziri, Nouha and Prabhumoye, Shrimai and Yang, Yiming and Gupta, Shashank and Majumder, Bodhisattwa Prasad and Hermann, Katherine and Welleck, Sean and Yazdanbakhsh, Amir and Clark, Peter},
  year = 2023,
  month = nov,
  url = {https://openreview.net/forum?id=S37hOerQLB},
  urldate = {2026-04-27},
  langid = {english}
}

@inproceedings{shinnReflexionLanguageAgents2023c,
  title = {Reflexion: Language Agents with Verbal Reinforcement Learning},
  shorttitle = {Reflexion},
  booktitle = {Thirty-Seventh Conference on Neural Information Processing Systems},
  author = {Shinn, Noah and Cassano, Federico and Gopinath, Ashwin and Narasimhan, Karthik R. and Yao, Shunyu},
  year = 2023,
  month = nov,
  url = {https://openreview.net/forum?id=vAElhFcKW6},
  urldate = {2026-04-27},
  langid = {english}
}

@inproceedings{zhangAgenticContextEngineering2025a,
  title = {Agentic Context Engineering: Evolving Contexts for Self-Improving Language Models},
  shorttitle = {Agentic Context Engineering},
  booktitle = {The Fourteenth International Conference on Learning Representations},
  author = {Zhang, Qizheng and Hu, Changran and Upasani, Shubhangi and Ma, Boyuan and Hong, Fenglu and Kamanuru, Vamsidhar and Rainton, Jay and Wu, Chen and Ji, Mengmeng and Li, Hanchen and Thakker, Urmish and Zou, James and Olukotun, Kunle},
  year = 2026,
  url = {https://openreview.net/forum?id=eC4ygDs02R},
  urldate = {2026-04-27},
  langid = {english}
}

@misc{wangVoyagerOpenEndedEmbodied2023a,
  title = {Voyager: An Open-Ended Embodied Agent with Large Language Models},
  shorttitle = {Voyager},
  author = {Wang, Guanzhi and Xie, Yuqi and Jiang, Yunfan and Mandlekar, Ajay and Xiao, Chaowei and Zhu, Yuke and Fan, Linxi and Anandkumar, Anima},
  year = 2023,
  month = oct,
  number = {arXiv:2305.16291},
  eprint = {2305.16291},
  primaryclass = {cs},
  publisher = {arXiv},
  doi = {10.48550/arXiv.2305.16291},
  url = {http://arxiv.org/abs/2305.16291},
  urldate = {2026-04-27},
  archiveprefix = {arXiv}
}

@misc{novikovAlphaEvolveCodingAgent2025a,
  title = {AlphaEvolve: A Coding Agent for Scientific and Algorithmic Discovery},
  shorttitle = {AlphaEvolve},
  author = {Novikov, Alexander and V{\~u}, Ng{\^a}n and Eisenberger, Marvin and Dupont, Emilien and Huang, Po-Sen and Wagner, Adam Zsolt and Shirobokov, Sergey and Kozlovskii, Borislav and Ruiz, Francisco J. R. and Mehrabian, Abbas and Kumar, M. Pawan and See, Abigail and Chaudhuri, Swarat and Holland, George and Davies, Alex and Nowozin, Sebastian and Kohli, Pushmeet and Balog, Matej},
  year = 2025,
  month = jun,
  number = {arXiv:2506.13131},
  eprint = {2506.13131},
  primaryclass = {cs},
  publisher = {arXiv},
  doi = {10.48550/arXiv.2506.13131},
  url = {http://arxiv.org/abs/2506.13131},
  urldate = {2026-04-27},
  archiveprefix = {arXiv}
}

@inproceedings{huAutomatedDesignAgentic2024,
  title = {Automated Design of Agentic Systems},
  booktitle = {The Thirteenth International Conference on Learning Representations},
  author = {Hu, Shengran and Lu, Cong and Clune, Jeff},
  year = 2025,
  url = {https://openreview.net/forum?id=t9U3LW7JVX},
  urldate = {2026-04-27},
  langid = {english}
}

@inproceedings{zhangAFlowAutomatingAgentic2024,
  title = {AFlow: Automating Agentic Workflow Generation},
  shorttitle = {AFlow},
  booktitle = {The Thirteenth International Conference on Learning Representations},
  author = {Zhang, Jiayi and Xiang, Jinyu and Yu, Zhaoyang and Teng, Fengwei and Chen, Xiong-Hui and Chen, Jiaqi and Zhuge, Mingchen and Cheng, Xin and Hong, Sirui and Wang, Jinlin and Zheng, Bingnan and Liu, Bang and Luo, Yuyu and Wu, Chenglin},
  year = 2025,
  url = {https://openreview.net/forum?id=z5uVAKwmjf},
  urldate = {2026-04-27},
  langid = {english}
}

@misc{zhang2026darwingodelmachineopenended,
      title={Darwin Godel Machine: Open-Ended Evolution of Self-Improving Agents}, 
      author={Jenny Zhang and Shengran Hu and Cong Lu and Robert Lange and Jeff Clune},
      year={2026},
      eprint={2505.22954},
      archivePrefix={arXiv},
      primaryClass={cs.AI},
      url={https://arxiv.org/abs/2505.22954}, 
}

@misc{lou2026autoharnessimprovingllmagents,
      title={AutoHarness: improving LLM agents by automatically synthesizing a code harness}, 
      author={Xinghua Lou and Miguel Lázaro-Gredilla and Antoine Dedieu and Carter Wendelken and Wolfgang Lehrach and Kevin P. Murphy},
      year={2026},
      eprint={2603.03329},
      archivePrefix={arXiv},
      primaryClass={cs.CL},
      url={https://arxiv.org/abs/2603.03329}, 
}

@misc{leeMetaHarnessEndtoEndOptimization2026,
  title = {Meta-Harness: End-to-End Optimization of Model Harnesses},
  shorttitle = {Meta-Harness},
  author = {Lee, Yoonho and Nair, Roshen and Zhang, Qizheng and Lee, Kangwook and Khattab, Omar and Finn, Chelsea},
  year = 2026,
  month = mar,
  number = {arXiv:2603.28052},
  eprint = {2603.28052},
  primaryclass = {cs},
  publisher = {arXiv},
  doi = {10.48550/arXiv.2603.28052},
  url = {http://arxiv.org/abs/2603.28052},
  urldate = {2026-04-15},
  archiveprefix = {arXiv}
}

@misc{linAgenticHarnessEngineering2026,
  title = {Agentic Harness Engineering: Observability-Driven Automatic Evolution of Coding-Agent Harnesses},
  shorttitle = {Agentic Harness Engineering},
  author = {Lin, Jiahang and Liu, Shichun and Pan, Chengjun and Lin, Lizhi and Dou, Shihan and Huang, Xuanjing and Yan, Hang and Han, Zhenhua and Gui, Tao},
  year = 2026,
  month = apr,
  number = {arXiv:2604.25850},
  eprint = {2604.25850},
  primaryclass = {cs},
  publisher = {arXiv},
  doi = {10.48550/arXiv.2604.25850},
  url = {http://arxiv.org/abs/2604.25850},
  urldate = {2026-04-29},
  archiveprefix = {arXiv}
}

@misc{robeyns2025selfimprovingcodingagent,
  title = {A Self-Improving Coding Agent},
  author = {Robeyns, Maxime and Szummer, Martin and Aitchison, Laurence},
  year = {2025},
  eprint = {2504.15228},
  archivePrefix = {arXiv},
  primaryClass = {cs.AI},
  doi = {10.48550/arXiv.2504.15228},
  url = {https://arxiv.org/abs/2504.15228}
}

@misc{wang2025huxleygodelmachine,
  title = {Huxley-Godel Machine: Human-Level Coding Agent Development by an Approximation of the Optimal Self-Improving Machine},
  author = {Wang, Wenyi and Piekos, Piotr and Li, Nanbo and Laakom, Firas and Chen, Yimeng and Ostaszewski, Mateusz and Zhuge, Mingchen and Schmidhuber, Jurgen},
  year = {2025},
  eprint = {2510.21614},
  archivePrefix = {arXiv},
  primaryClass = {cs.AI},
  doi = {10.48550/arXiv.2510.21614},
  url = {https://arxiv.org/abs/2510.21614}
}

@misc{xia2025livesweagent,
  title = {Live-SWE-agent: Can Software Engineering Agents Self-Evolve on the Fly?},
  author = {Xia, Chunqiu Steven and Wang, Zhe and Yang, Yan and Wei, Yuxiang and Zhang, Lingming},
  year = {2025},
  eprint = {2511.13646},
  archivePrefix = {arXiv},
  primaryClass = {cs.SE},
  doi = {10.48550/arXiv.2511.13646},
  url = {https://arxiv.org/abs/2511.13646}
}

@article{zhaoExpeLLLMAgents2024,
  title = {ExpeL: LLM Agents Are Experiential Learners},
  shorttitle = {ExpeL},
  author = {Zhao, Andrew and Huang, Daniel and Xu, Quentin and Lin, Matthieu and Liu, Yong-Jin and Huang, Gao},
  journal = {Proceedings of the AAAI Conference on Artificial Intelligence},
  volume = {38},
  number = {17},
  pages = {19632--19642},
  year = {2024},
  doi = {10.1609/aaai.v38i17.29936},
  url = {https://ojs.aaai.org/index.php/AAAI/article/view/29936}
}

@article{ye2026meta,
  title = {Meta Context Engineering via Agentic Skill Evolution},
  author = {Ye, Haoran and He, Xuning and Arak, Vincent and Dong, Haonan and Song, Guojie},
  journal = {arXiv preprint arXiv:2601.21557},
  year = {2026}
}

@misc{maSkillClawLetSkills2026b,
  title = {SkillClaw: Let Skills Evolve Collectively with Agentic Evolver},
  shorttitle = {SkillClaw},
  author = {Ma, Ziyu and Yang, Shidong and Ji, Yuxiang and Wang, Xucong and Wang, Yong and Hu, Yiming and Huang, Tongwen and Chu, Xiangxiang},
  year = 2026,
  month = apr,
  number = {arXiv:2604.08377},
  eprint = {2604.08377},
  primaryclass = {cs},
  publisher = {arXiv},
  doi = {10.48550/arXiv.2604.08377},
  url = {http://arxiv.org/abs/2604.08377},
  urldate = {2026-04-30},
  archiveprefix = {arXiv}
}

@misc{xiaSkillRLEvolvingAgents2026c,
  title = {SkillRL: Evolving Agents via Recursive Skill-Augmented Reinforcement Learning},
  shorttitle = {SkillRL},
  author = {Xia, Peng and Chen, Jianwen and Wang, Hanyang and Liu, Jiaqi and Zeng, Kaide and Wang, Yu and Han, Siwei and Zhou, Yiyang and Zhao, Xujiang and Chen, Haifeng and Zheng, Zeyu and Xie, Cihang and Yao, Huaxiu},
  year = 2026,
  month = feb,
  number = {arXiv:2602.08234},
  eprint = {2602.08234},
  primaryclass = {cs},
  publisher = {arXiv},
  doi = {10.48550/arXiv.2602.08234},
  url = {http://arxiv.org/abs/2602.08234},
  urldate = {2026-04-30},
  archiveprefix = {arXiv}
}

@misc{guertler2025textarena,
  title = {TextArena},
  author = {Guertler, Leon and Cheng, Bobby and Yu, Simon and Liu, Bo and Choshen, Leshem and Tan, Cheston},
  year = {2025},
  eprint = {2504.11442},
  archivePrefix = {arXiv},
  primaryClass = {cs.CL},
  doi = {10.48550/arXiv.2504.11442},
  url = {https://arxiv.org/abs/2504.11442}
}

@misc{paglieri2024balrog,
  title = {BALROG: Benchmarking Agentic LLM and VLM Reasoning On Games},
  author = {Paglieri, Davide and Cupial, Bartlomiej and Coward, Samuel and Piterbarg, Ulyana and Wolczyk, Maciej and Khan, Akbir and Pignatelli, Eduardo and Kucinski, Lukasz and Pinto, Lerrel and Fergus, Rob and Foerster, Jakob Nicolaus and Parker-Holder, Jack and Rocktaschel, Tim},
  year = {2024},
  eprint = {2411.13543},
  archivePrefix = {arXiv},
  primaryClass = {cs.AI},
  doi = {10.48550/arXiv.2411.13543},
  url = {https://arxiv.org/abs/2411.13543}
}

@misc{li2025textatari,
  title = {TextAtari: 100K Frames Game Playing with Language Agents},
  author = {Li, Wenhao and Li, Wenwu and Shen, Chuyun and Sheng, Junjie and Huang, Zixiao and Wu, Di and Hua, Yun and Yin, Wei and Wang, Xiangfeng and Zha, Hongyuan and Jin, Bo},
  year = {2025},
  eprint = {2506.04098},
  archivePrefix = {arXiv},
  primaryClass = {cs.CL},
  doi = {10.48550/arXiv.2506.04098},
  url = {https://arxiv.org/abs/2506.04098}
}

@misc{hu2024gamearena,
  title = {GameArena: Evaluating LLM Reasoning through Live Computer Games},
  author = {Hu, Lanxiang and Li, Qiyu and Xie, Anze and Jiang, Nan and Stoica, Ion and Jin, Haojian and Zhang, Hao},
  year = {2024},
  eprint = {2412.06394},
  archivePrefix = {arXiv},
  primaryClass = {cs.AI},
  doi = {10.48550/arXiv.2412.06394},
  url = {https://arxiv.org/abs/2412.06394}
}

@misc{duan2024gtbench,
  title = {GTBench: Uncovering the Strategic Reasoning Limitations of LLMs via Game-Theoretic Evaluations},
  author = {Duan, Jinhao and Zhang, Renming and Diffenderfer, James and Kailkhura, Bhavya and Sun, Lichao and Stengel-Eskin, Elias and Bansal, Mohit and Chen, Tianlong and Xu, Kaidi},
  year = {2024},
  eprint = {2402.12348},
  archivePrefix = {arXiv},
  primaryClass = {cs.CL},
  doi = {10.48550/arXiv.2402.12348},
  url = {https://arxiv.org/abs/2402.12348}
}

@misc{park2025orak,
  title = {Orak: A Foundational Benchmark for Training and Evaluating LLM Agents on Diverse Video Games},
  author = {Park, Dongmin and Kim, Minkyu and Choi, Beongjun and Kim, Junhyuck and Lee, Keon and Lee, Jonghyun and Park, Inkyu and Lee, Byeong-Uk and Hwang, Jaeyoung and Ahn, Jaewoo and Mahabaleshwarkar, Ameya S. and Kartal, Bilal and Biswas, Pritam and Suhara, Yoshi and Lee, Kangwook and Cho, Jaewoong},
  year = {2025},
  eprint = {2506.03610},
  archivePrefix = {arXiv},
  primaryClass = {cs.AI},
  doi = {10.48550/arXiv.2506.03610},
  url = {https://arxiv.org/abs/2506.03610}
}

@misc{fu2026catarenaevaluatingevolutionarycapabilities,
  title={CATArena: Evaluating Evolutionary Capabilities of Code Agents via Iterative Tournaments},
  author={Lingyue Fu and Xin Ding and Linyue Pan and Yaoming Zhu and Shao Zhang and Lin Qiu and Weiwen Liu and Weinan Zhang and Xuezhi Cao and Xunliang Cai and Jiaxin Ding and Yong Yu},
  year={2026},
  eprint={2510.26852},
  archivePrefix={arXiv},
  primaryClass={cs.AI},
  url={https://arxiv.org/abs/2510.26852},
}

@inproceedings{yao2022react,
  title = {ReAct: Synergizing Reasoning and Acting in Language Models},
  author = {Yao, Shunyu and Zhao, Jeffrey and Yu, Dian and Du, Nan and Shafran, Izhak and Narasimhan, Karthik R. and Cao, Yuan},
  booktitle = {The Eleventh International Conference on Learning Representations},
  year = {2023},
  url = {https://openreview.net/forum?id=WE_vluYUL-X}
}

@misc{coder2026balatrollm,
  title = {{BalatroLLM}: Play {Balatro} with {LLMs}},
  author = {{Coder}},
  year = {2026},
  howpublished = {\url{https://github.com/coder/balatrollm}},
  note = {Accessed May 2, 2026}
}

@misc{coder2026balatrobench,
  title = {BalatroBench},
  author = {{Coder}},
  year = {2026},
  howpublished = {\url{https://balatrobench.com/}},
  note = {Accessed May 2, 2026}
}

@inproceedings{ross2011dagger,
  title = {A Reduction of Imitation Learning and Structured Prediction to No-Regret Online Learning},
  author = {Ross, St{\'e}phane and Gordon, Geoffrey and Bagnell, Drew},
  booktitle = {Proceedings of the Fourteenth International Conference on Artificial Intelligence and Statistics},
  pages = {627--635},
  year = {2011}
}

@inproceedings{ho2016generative,
  title = {Generative Adversarial Imitation Learning},
  author = {Ho, Jonathan and Ermon, Stefano},
  booktitle = {Advances in Neural Information Processing Systems},
  year = {2016}
}

@article{hester2018deep,
  title = {Deep Q-learning from Demonstrations},
  author = {Hester, Todd and Vecerik, Matej and Pietquin, Olivier and Lanctot, Marc and Schaul, Tom and Piot, Bilal and Horgan, Dan and Quan, John and Sendonaris, Andrew and Osband, Ian and Dulac-Arnold, Gabriel and Agapiou, John and Leibo, Joel Z. and Gruslys, Audrunas},
  journal = {Proceedings of the AAAI Conference on Artificial Intelligence},
  volume = {32},
  number = {1},
  year = {2018}
}

@inproceedings{nair2018overcoming,
  title = {Overcoming Exploration in Reinforcement Learning with Demonstrations},
  author = {Nair, Ashvin and McGrew, Bob and Andrychowicz, Marcin and Zaremba, Wojciech and Abbeel, Pieter},
  booktitle = {2018 IEEE International Conference on Robotics and Automation (ICRA)},
  pages = {6292--6299},
  year = {2018}
}

@inproceedings{rajeswaran2018learning,
  title = {Learning Complex Dexterous Manipulation with Deep Reinforcement Learning and Demonstrations},
  author = {Rajeswaran, Aravind and Kumar, Vikash and Gupta, Abhishek and Vezzani, Giulia and Schulman, John and Todorov, Emanuel and Levine, Sergey},
  booktitle = {Robotics: Science and Systems},
  year = {2018}
}

@incollection{bengio2009curriculum,
  title = {Curriculum Learning},
  author = {Bengio, Yoshua and Louradour, J{\'e}r{\^o}me and Collobert, Ronan and Weston, Jason},
  booktitle = {Proceedings of the 26th Annual International Conference on Machine Learning},
  pages = {41--48},
  publisher = {Association for Computing Machinery},
  year = {2009}
}

@article{kumar2010selfpaced,
  title = {Self-Paced Learning for Latent Variable Models},
  author = {Kumar, M. Pawan and Packer, Benjamin and Koller, Daphne},
  journal = {Advances in Neural Information Processing Systems},
  volume = {23},
  year = {2010}
}

@inproceedings{florensa2017reverse,
  title = {Reverse Curriculum Generation for Reinforcement Learning},
  author = {Florensa, Carlos and Held, David and Wulfmeier, Markus and Zhang, Michael and Abbeel, Pieter},
  booktitle = {Conference on Robot Learning},
  pages = {482--495},
  year = {2017}
}

@article{ecoffet2021first,
  title = {First Return, Then Explore},
  author = {Ecoffet, Adrien and Huizinga, Joost and Lehman, Joel and Stanley, Kenneth O. and Clune, Jeff},
  journal = {Nature},
  volume = {590},
  number = {7847},
  pages = {580--586},
  year = {2021}
}

@misc{chartyr2026sts2agent,
  title = {{STS2-Agent}: A Mod and {MCP} Server for {Slay the Spire 2}},
  author = {{CharTyr}},
  year = {2026},
  howpublished = {\url{https://github.com/CharTyr/STS2-Agent}},
  note = {Accessed July 31, 2026}
}

@misc{coder2026balatrobot,
  title = {{BalatroBot}: An {API} for Developing {Balatro} Bots},
  author = {{Coder}},
  year = {2026},
  howpublished = {\url{https://github.com/coder/balatrobot}},
  note = {Accessed July 31, 2026}
}

\clearpage
\appendix
\setcounter{secnumdepth}{2}
% Appendix entry point. Individual appendix sections live in src/appendix/.
\section{TextArena Liar's Dice Details}
\label{app:textarena-reproducibility}

This appendix gives additional details for the TextArena Liar's Dice experiments
reported in the main paper~\citep{guertler2025textarena}. These experiments serve as a
lightweight positive control: they test whether self-rollout harness evolution
works when feedback is relatively short-horizon and attributable, before moving
to the longer-horizon Balatro and STS2 settings.

\paragraph{Game variants.}
We report two two-player Liar's Dice variants. \textbf{Small3} uses the
\texttt{LiarsDice-v0-\allowbreak small} environment: each player starts with three dice,
ones are not wild, players alternately bid a quantity/face pair or call the
previous bid, and losing a challenge removes one die. \textbf{OneCall-Wild1}
uses the \texttt{LiarsDice-v0-\allowbreak onecall} environment: each player starts with five
dice, ones are wild until face 1 is bid, and the first \texttt{[Call]}
immediately ends the match without dice loss or rerolls. These variants are
interactive and stochastic, with hidden dice, opponent pressure, legal-action
constraints, and sequential bidding decisions. However, compared with Balatro,
each episode is short enough that terminal outcomes can usually be traced back
to a small number of bid or call decisions. They therefore provide a cleaner
setting for testing whether self-generated rollout traces can support harness
edits.

\paragraph{Role as a lightweight positive control.}
We use TextArena Liar's Dice to check that the outer-loop machinery can improve
a frozen-model agent when failures are attributable and the search signal is
relatively dense. The point of this experiment is not to show that self-rollout
feedback is sufficient for sparse, long-horizon games. Instead, it establishes a
positive case: when the task horizon is short and rollout failures are easy to
localize, self-rollout harness evolution can convert observed failures into
useful executable harness structure.

\paragraph{Search and selection.}
For each target/opponent/task cell, the outer-loop proposer modifies only the
external harness around a frozen target model. Candidate selection uses
development/search rollouts only; held-out seeds are not visible during candidate
proposal or selection. The search budget is deliberately small: candidate
evaluations usually use \(N=5\) logical development seeds, while some later
Small3 candidates and baselines use \(N=10\). The selected harness is chosen
according to development reward only. One OneCall-Wild1 GPT self-play run had a
pre-declared development-only tie-break: two candidates both reached development
reward \(0.700\), and the clean tie-break selected the later candidate. This
tie-break did not use held-out performance.

\paragraph{Held-out paired evaluation.}
Final evaluation uses 30 matched logical held-out seeds for every base/evolved
comparison. Small3 uses held-out logical seeds \(120000\)--\(120029\), and
OneCall-Wild1 uses held-out logical seeds \(130000\)--\(130029\). The same seed
set is used for the base harness and the selected evolved harness within each
cell.

We run held-out evaluation with paired-side control. Each logical seed is
evaluated as two raw games with the same environment seed and swapped player
slots. The reported paired reward averages the two legs, so a win receives
reward 1, a draw receives reward 0.5, and a loss receives reward 0. Thus
\(N=30\) means 30 paired logical matches, or 60 raw games, for each base or
evolved row. This controls for first-player and seat-order effects while keeping
the stochastic dice rolls fixed across the base and evolved comparisons.

\paragraph{Metrics and clean-run criteria.}
The primary metric is mean held-out paired reward over the 30 logical seeds,
with standard errors computed across paired rewards. The paper-level aggregate
is task-balanced: we average run-level means within Small3 and OneCall-Wild1,
then average the two task means. In the clean tie-break package, this gives
\(0.392\) for the base harness and \(0.800\) for the selected evolved harness,
an absolute gain of \(+0.408\).

A row is included in the clean main result only if all 30 base and 30 evolved
paired summaries are present, paired-side evaluation succeeds for all logical
seeds, and there are no LLM fallback, error, or timeout events on either target
or opponent side. Invalid actions and parse fallbacks are logged as behavioral
audit metrics rather than infrastructure failures.

\begin{figure}[t]
\centering
\includegraphics[width=\linewidth]{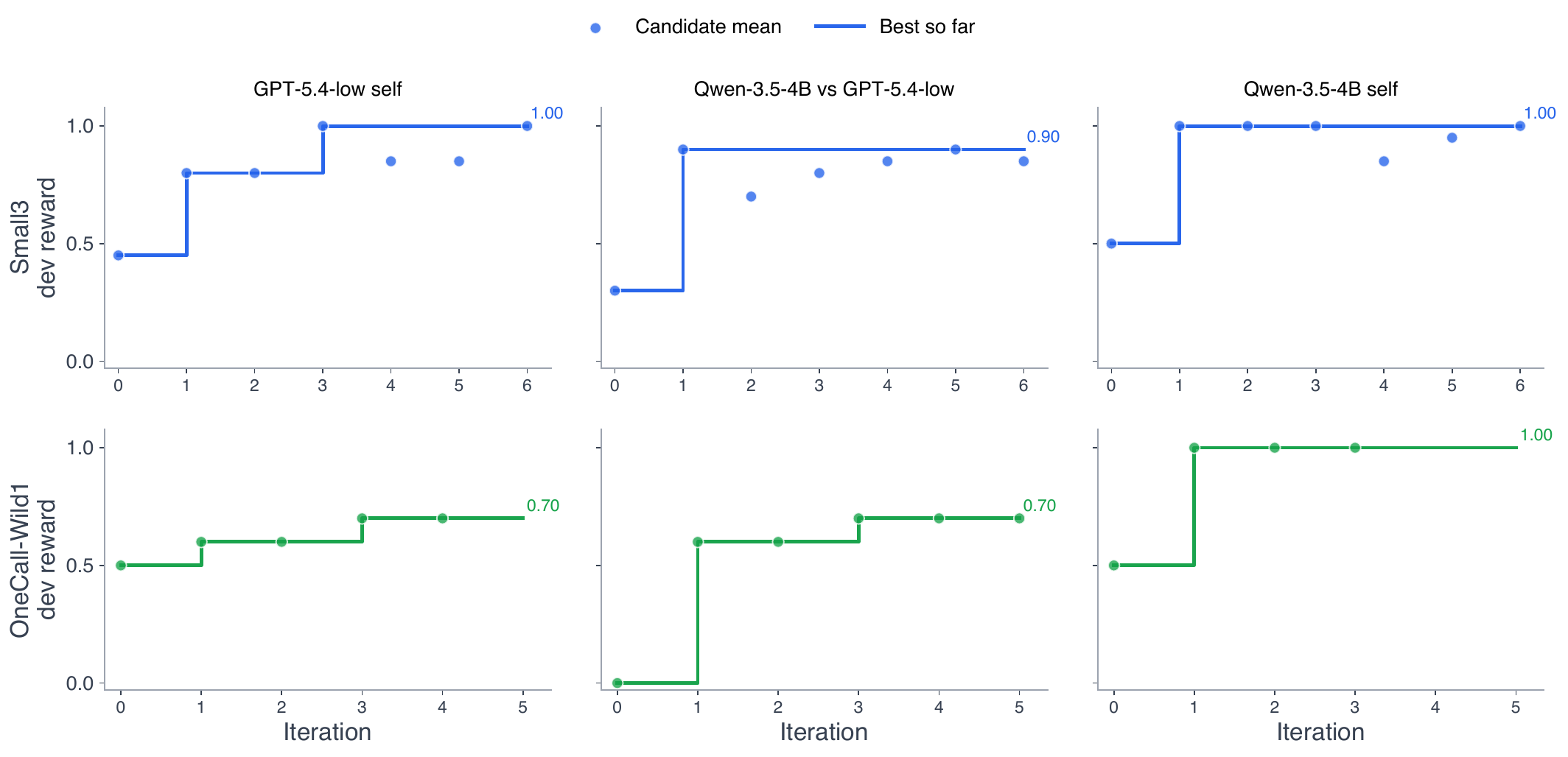}
\caption{
Development/search progress for TextArena Liar's Dice. Iteration 0 is the base
harness. Points show candidate development reward, and lines show the best
development reward observed so far. Candidate selection uses only
development/search rollouts; held-out seeds are not used during search.
}
\label{fig:textarena-search-progress-app}
\end{figure}

\paragraph{Artifact regeneration.}
The reported TextArena numbers are generated from archived clean tie-break
artifacts containing raw held-out traces, held-out summaries, macro summaries,
audit files, source snapshots, and per-candidate search-progress records.
Execution parallelism only controls concurrency and does not repeat evaluation
on the same seed. As with any evaluation using hosted or locally served LLMs,
exact bit-for-bit replay can depend on model-serving implementation, endpoint
availability, and retry behavior. The reported numbers are therefore tied to the
archived raw traces and summaries, while the seeded protocol specifies the
stochastic game instances used for reproduction.

\section{Qualitative Analysis of Evolved Liar's Dice Harnesses}
\label{app:liarsdice_qualitative}

This section qualitatively inspects selected Liar's Dice harnesses from the
TextArena experiments in the main paper. The goal is not to present the full source
code, but to understand what kinds of executable heuristics self-rollout harness
evolution discovered. Across the selected harnesses, we observe a recurring
pattern: the outer loop moves brittle decision computations out of free-form
language-model reasoning and into executable scaffolding. Depending on the base
model and game variant, this scaffolding ranges from near-symbolic control to a
neuro-symbolic division of labor.

\paragraph{Case A: Near-symbolic threshold control for a weaker model.}
In the Small3 Qwen self-play setting, the selected harness largely bypasses the
language model at decision time. It parses the observation, estimates whether
the current bid is true, calls when the bid appears sufficiently unlikely, and
otherwise raises to a minimally legal bid with acceptable probability. A simplified
sketch is:

\begin{footnotesize}
\begin{verbatim}
state = parse(observation)

if no_bid:
    return opening_bid(state)

p_true = claim_probability(
    current_bid, own_dice, opp_dice)

if p_true < 0.30:
    return "[Call]"

raise = smallest_legal_raise(
    min_probability=0.50)
return raise
\end{verbatim}
\end{footnotesize}

This case is useful precisely because it is close to a limiting case. Harness
evolution is not constrained to preserve a fixed role for the language model.
When the base model is weak or unstable for the local game mechanics, the search
can improve the overall frozen-model agent by shifting more of the decision
procedure into deterministic code. In this lightweight setting, the resulting
``skill'' resembles a compact rule-of-thumb policy rather than improved model
reasoning: the optimized object is the full model--harness system.

\paragraph{Case B: Opponent-aware expected-value scoring.}
In the Small3 GPT-5.4 self-play setting, the selected harness keeps a more hybrid
structure. The harness parses the accumulated observation history, constructs a
simple opponent face prior from the opponent's bids in the current round, computes
binomial-tail probabilities for the current claim and candidate raises, and ranks
legal actions by expected value. A simplified sketch is:

\begin{footnotesize}
\begin{verbatim}
opp_prior = opponent_face_prior(
    bids_this_round)

p_current = claim_probability(
    current_bid, own_dice,
    opp_dice_count, opp_prior)
ev_call = 2 * (1 - p_current) - 1

for raise in legal_raises(current_bid):
    p_raise = claim_probability(
        raise, own_dice,
        opp_dice_count, opp_prior)
    ev_raise = 2 * p_raise - 1

return best_by_ev(call, legal_raises)
\end{verbatim}
\end{footnotesize}

This harness does not merely decide what text to show the model. It computes
decision-relevant variables that the model is brittle at maintaining across
turns: legal raises, claim probabilities, opponent-conditioned priors, and
call-versus-raise tradeoffs. The language model can still be used for residual
strategic judgment or action verbalization, but the fragile arithmetic and
legality constraints are stabilized outside the model.

\paragraph{Case C: Bayesian pile-on guard in OneCall-Wild1.}
OneCall-Wild1 is a harder Liar's Dice variant because ones are wild and a single
call ends the match. In the selected GPT-5.4 harness, the evolved logic adds a
state-conditioned guard for deep same-face escalation. The harness estimates the
truth probability of the current bid, estimates the best follow-up raise under a
Bayesian pile-on model conditioned on the opponent's last bid, and overrides
continued raising when calling has higher expected value by a margin:

\begin{footnotesize}
\begin{verbatim}
p_truth = claim_probability(
    current_bid, own_dice, wild_rule)
best_raise = argmax_raise(
    opponent_last_bid, wild_rule)

ev_call = 1.0 - p_truth
ev_raise = raise_probability(
    best_raise, pile_on_model)

if own_contribution >= 2 and
   ev_call > ev_raise + margin:
    action = "[Call]"
else:
    action = best_raise
\end{verbatim}
\end{footnotesize}

This case illustrates a more explicitly neuro-symbolic pattern. The harness
handles rule interpretation, probability estimation, legal-action validation,
and a high-risk override, while still allowing model-facing analysis or
recommendations in ambiguous states. The intervention is state-conditioned rather
than seed-conditioned: it fires on a structural pattern in the bidding trajectory
rather than on a particular development seed.

\paragraph{Takeaway.}
The Liar's Dice case studies show that self-rollout harness evolution can
discover compact executable game heuristics. In easier or weaker-model settings,
the selected harness may become nearly symbolic, replacing the model's unstable
local decisions with deterministic thresholds and guards. In harder variants or
with stronger models, the harness tends to form a hybrid system: symbolic code
handles explicit rules, probabilities, legal actions, and safety constraints,
while the frozen language model handles residual semantic and strategic judgment.
This supports the view that harness evolution optimizes the behavior of the
entire model--harness system, not merely the prompt or the model's standalone
generation quality.

\section{Shared Long-Horizon Experimental Protocol}
\label{app:shared-long-horizon-protocol}
\label{app:experimental-details}

Balatro and STS2 use the same evolution and evaluation design. Within each environment,
three of five fixed seeds are used for development and the remaining two are held out. The
proposer never observes held-out rollouts, scores, or traces during evolution. Candidate
harnesses are evaluated with one rollout on each development seed during search. After
development-only selection, the final harness is evaluated from scratch with three
independent rollouts on all five seeds.

All evolution conditions run for 10 outer-loop iterations with GPT-5.4 at low reasoning
effort as the frozen task-execution model and Claude Opus 4.7 at maximum reasoning effort
as the coding proposer. Within each
environment, the initial harness, search budget, proposer, candidate-selection rule, and final
evaluation protocol are held fixed. \textbf{Meta-Harness} uses prior harness code,
development scores, self-generated rollouts, and execution traces. \textbf{Meta-Harness-Web}
additionally gives the proposer web-search access but no human trajectories.
\textbf{DemoEvolve} instead provides nine competent human trajectories collected on the
three development seeds. These trajectories are available only to the proposer and are never
inserted into the task-execution model's context during final evaluation.

\subsection{Compute and API Cost}
\label{sec:appendix-long-horizon-cost}

API usage profiles differ across the two environments because STS2 rollout length varies
with progress. For Balatro, the 15-rollout final evaluation costs approximately \$32, or \$2.13 per
rollout. A typical 10-iteration evolution run evaluates about two candidates per iteration on
three development seeds, giving approximately \(10\times2\times3=60\) development rollouts at an
estimated cost of \$128. Including final evaluation, the Balatro task-execution cost is about
\$160 per evolution run.

STS2 cost varies with survival depth. Across measured runs, the median rollout makes 211
model calls and consumes 2.2M input tokens. The corresponding medians are 334 calls and
3.4M tokens for Act~1 clears, 160 calls and 1.6M tokens for unsuccessful runs, and 375 calls
and 4.4M tokens for runs reaching the Act~2 boss. Output usage is comparatively small
because each response contains only an action identifier and a one-line rationale. Under our
API cost accounting, the 15-rollout STS2 final evaluation costs approximately \$34. We do
not extrapolate a fixed STS2 evolution-run total because candidate cost depends strongly on
progress. All estimates include task-execution calls only
and exclude the coding proposer.
\section{Human Demonstration Collection}
\label{sec:appendix-human-demos}

Experts played through the native game interfaces while a game-side mod streamed live
state to a recorder. We used neither scripted play nor manually entered action labels.
Instead, the recorder repeatedly polled the game state, inferred the player action associated
with each observed state transition, and stored the result using the same observation--action
schema used for agent rollouts. Demonstrations and
agent rollouts can therefore be compared in a common representation without action-level
manual annotation.

For Balatro, we recorded through the BalatroBot mod~\citep{coder2026balatrobot}, which exposes
the running game through a JSON-RPC interface. The recorder polled the state several times
per second and derived actions from successive state changes. For STS2, we implemented a
headed recording mod adapted from STS2-Agent~\citep{chartyr2026sts2agent} that serves the
same role.

Before using the trajectories, we remove two classes of artifacts introduced by live-state
recording: automatic state transitions that do not correspond to player decisions, such as
card draws and automatic resolutions, and interface-only interactions with no game effect,
such as zero-effect selections or opening and closing an inspection view without acting.

The Balatro corpus contains nine demonstration runs meeting the benchmark success criterion
on the three development seeds, with three
runs per seed and a mean of approximately 342 recorded decisions per run (3,080 total).
Sessions lasted 27--38 minutes. The STS2 corpus likewise contains nine demonstration runs
meeting its benchmark success criterion, three per development seed, with approximately
390 decisions per run
(approximately 3,510 total). Sessions lasted 39--55 minutes, with a median of 44 minutes.

Players could optionally enter commentary at any point. This produced 59 verbatim
rationales in Balatro and 72 in STS2. These comments are the only free-text supervision in
the demonstration corpora. They are stored unchanged with the corresponding trajectories
and supplied verbatim to the coding proposer or, for the demonstrations-without-evolution
baselines, to the summary-generation prompts.

\FloatBarrier
\section{Balatro Background and Terminology}
\label{sec:appendix}
\label{sec:appendix-balatro-background}

\subsection{Game Rules and Terminology}
\label{sec:appendix-balatro-rules}
\begin{figure*}[t]
    \centering
    \includegraphics[width=0.82\textwidth]{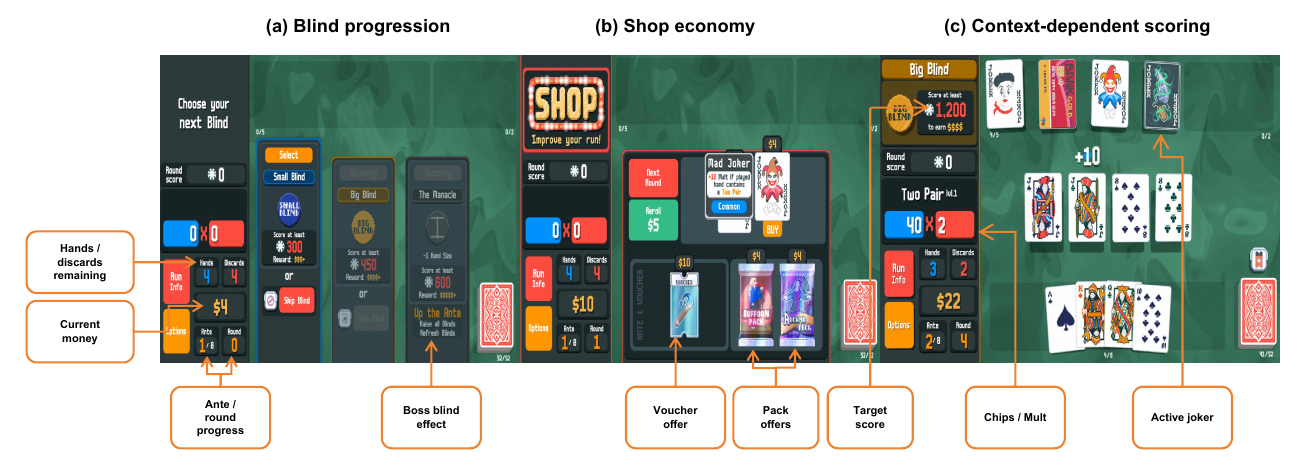}
    \caption{
    Annotated Balatro screenshots illustrating the mechanics used in our analysis.
    \textbf{(a)} Blind progression defines the survival objective: the agent must clear increasing target scores with limited hands and discards, while boss blinds add special constraints.
    \textbf{(b)} The shop phase turns money into a long-horizon resource, since the agent must choose among purchases, rerolls, and saving for future shops.
    \textbf{(c)} Scoring is build-dependent: the same played hand can have different value depending on chips, multipliers, remaining resources, and active jokers.
    }
    \label{fig:balatro_mechanics}
\end{figure*}

This section provides a compact introduction to Balatro for readers who have not played the game. We focus on the mechanics needed to interpret our benchmark, metrics, and behavioral analyses.

\paragraph{Run structure.}
A Balatro run consists of a sequence of \emph{antes}, each containing several \emph{blinds}. To clear a blind, the player must reach a target score before exhausting the available hands and discards. Failure to reach the target ends the run. We therefore use a rollout's \emph{final round} as a coarse measure of survival depth.

\paragraph{Scoring and hand play.}
During each blind, the player plays poker-style hands. The score of a played hand depends on the hand category, card values, hand levels, multipliers, and active modifiers. As a result, the best action is often not the strongest conventional poker hand in isolation; it depends on the current build, available scaling effects, and remaining resources.

\paragraph{Shops and economy.}
After blinds, the player enters a shop phase and can spend money on jokers, vouchers, booster packs, consumables, or rerolls. Money is therefore a strategic resource rather than a purely local reward signal. Spending aggressively may improve immediate survival, but preserving money can enable stronger future shops and more flexible scaling. This motivates our analysis of \emph{shop-entry money} as a long-horizon behavioral probe.

\paragraph{Jokers and synergies.}
Jokers are persistent modifiers that often define a run's strategy. Their value is highly compositional: a joker may be weak on its own, strong only with particular deck edits or consumables, or useful early but less valuable later. This context dependence makes action values difficult to interpret without considering the current joker set, deck composition, and stage of the run.

\paragraph{Randomness and seeds.}
Balatro contains substantial stochasticity, including shop inventories, pack contents, boss blind effects, and card draws. Fixing a seed makes a run reproducible, but repeated optimization on a small set of seeds can still exploit seed-specific opportunities rather than learn generally useful strategies. We therefore distinguish between development seeds used during harness evolution and held-out evaluation seeds used only for final evaluation.

\paragraph{Terminology used in the main text.}
The main paper uses several Balatro-specific terms:
\begin{itemize}
    \item \textbf{Blind}: a scoring challenge that must be cleared to continue the run.
    \item \textbf{Ante}: a stage of the run containing multiple blinds; later antes generally require higher scores.
    \item \textbf{Shop-entry money}: the amount of money available when entering a shop after a blind.
    \item \textbf{Joker set}: the current collection of jokers shaping the run's scoring, economy, and scaling behavior.
    \item \textbf{Boss blind}: a blind with an additional constraint or effect that can disrupt otherwise strong tactics.
    \item \textbf{Reroll}: spending money to refresh the shop's offerings.
\end{itemize}

% Declared before the next page begins so this double-column float can appear
% at the top of the page containing the STS2 background.
\begin{figure*}[t]
    \centering
    \IfFileExists{images/sts2_v01032_interfaces.png}{%
        \includegraphics[width=0.8\textwidth]{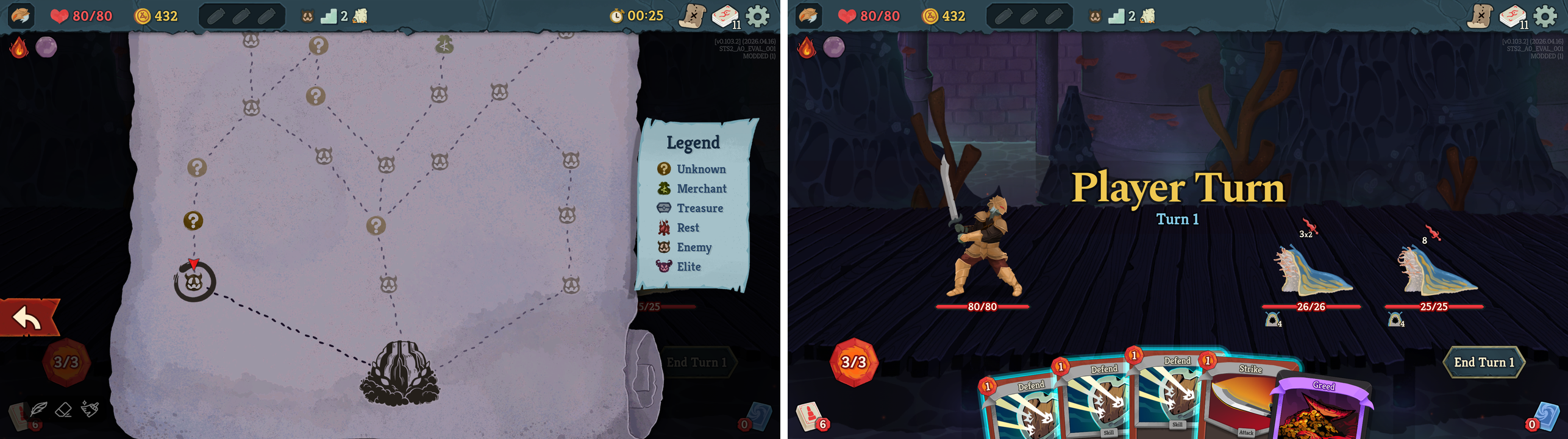}%
    }{%
        \fbox{\parbox[c][0.13\textheight][c]{0.82\linewidth}{\centering
        Missing image: \texttt{images/sts2\_v01032\_interfaces.png}}}%
    }
    \caption{Illustrative \emph{Slay the Spire 2} interfaces.
    \textbf{Left:} A run branches among combats, events, merchants, rest sites,
    and other encounters, requiring route choices that balance immediate risk
    against future resources. \textbf{Right:} In combat, the agent selects and
    orders cards using limited energy while accounting for its hand, health,
    and the enemy's announced intent. The screenshots and experiments use
    game version \texttt{v0.103.2}.}
    \label{fig:sts2-interfaces}
\end{figure*}

\subsection{Experimental Protocol}
\label{sec:appendix-balatro-protocol}

\paragraph{Game configuration.}
All Balatro experiments use game version \texttt{v1.0.1o-FULL}.

\paragraph{Environment-specific seeds.}
Following BalatroBench~\citep{coder2026balatrobench}, we use the literal seed strings
\texttt{AAAAAAA}--\texttt{EEEEEEE}, abbreviated as A--E. Seeds A/B/C form the
development panel and D/E form the held-out panel under the shared protocol in
Appendix~\ref{app:shared-long-horizon-protocol}.

\paragraph{Search and selection.}
All conditions start from the BalatroLLM harness~\citep{coder2026balatrollm}. Following the
Meta-Harness protocol~\citep{leeMetaHarnessEndtoEndOptimization2026}, the proposer may
inspect the evolution archive and choose which prior candidate to extend. We select the
harness with the highest mean final round on A/B/C, using completion rate as the tie-breaker
when necessary; all other search and evaluation settings follow the shared protocol.

\paragraph{Rationale for the fixed-seed low-sample protocol.}
BalatroBench was originally designed as a fixed-seed model-evaluation setting rather than
as a harness-evolution benchmark. We adopt fixed seeds because they make candidate
comparisons diagnosable during outer-loop search: if each candidate were evaluated on
different game instances, score changes would conflate harness edits with seed-level
variation. At the same time, evaluating every candidate on many seeds and rollouts would
make each evolution step slow and expensive, since every candidate requires full long-horizon
LLM-agent rollouts. Our protocol therefore uses a low-sample development budget during
search and a fixed-seed re-evaluation of the selected candidate. This is an engineering
trade-off rather than a claim that the resulting scores are variance-free estimates of average
Balatro skill. We interpret the fixed-seed results together with held-out-seed transfer,
functional audits, and behavioral analyses.
\section{Slay the Spire 2 Background}
\label{sec:appendix-sts2-background}

\paragraph{Game configuration.}
All STS2 experiments use game version \texttt{v0.103.2}, the
\emph{Ironclad} character, and Ascension~0 (A0). Our game-side mod is adapted
from the open-source STS2-Agent project~\citep{chartyr2026sts2agent}, which
exposes game state and actions through a local HTTP API. We modify the interface
to support the structured observation--action and evaluation protocol used by
STS2Bench.

\emph{Slay the Spire 2}~\citep{megacrit2026slaythespire2} is a turn-based
deck-building roguelike in which a run
alternates between selecting routes on a branching map and making node-level
decisions in combat, reward, shop, rest, and event states. Although each combat
is local, health, deck composition, relics, potions, and gold persist across
encounters, so card plays and resource decisions can affect survival many floors
later.

During combat, the agent observes its hand, available energy, status effects,
and enemy intents, then selects and orders card plays before ending the turn.
Combat outcomes affect persistent health and are followed by card-reward
decisions that modify the deck, linking turn-level tactics to subsequent
encounters.

\subsection{Experimental Protocol}
\label{sec:appendix-sts2-protocol}

\paragraph{Progress metric.}
A standard run spans three acts. The game numbers Act~1's 16 pre-boss map rows
followed by its boss as floor~17. We report cumulative floor across acts; hence,
17 means reaching the Act~1 boss, while Act~1 success requires defeating it.

\paragraph{Environment-specific seeds.}
The literal game-seed prefix is \texttt{STS2\_A0\_EVAL\_}: suffixes
\texttt{001}--\texttt{003} form the development panel and
\texttt{004}--\texttt{005} form the held-out panel. Their use otherwise follows the shared
protocol in Appendix~\ref{app:shared-long-horizon-protocol}. The \texttt{EVAL} component is
only a naming prefix and does not indicate that held-out seeds were exposed during evolution.

\FloatBarrier
\section{Functional Audit of the Selected Balatro Candidate}
\label{sec:appendix-inert-hook-failure}

This appendix audits the final candidate selected by the Meta-Harness
self-rollout baseline as its best Balatro harness under the A/B/C
development-seed selection rule. The selected candidate was
\texttt{face\_exposure\_shop\_hook}. It proposed a SHOP-phase hook for
face-card-dependent builds, intended to warn the agent before \textit{The Plant},
a boss blind that debuffs all face cards. The source contained a guarded prompt
block:
\begin{quote}
\begin{footnotesize}
\begin{verbatim}
{% if hooks.face_exposure %}
...
Face-Card Exposure Metrics
...
\end{verbatim}
\end{footnotesize}
\end{quote}
and candidate selection favored this version on the A/B/C development rollouts.

We audited the rendered model requests to test whether the proposed mechanism
actually reached the model. Across the three A/B/C validation rollouts, we
inspected 517 requests in \texttt{requests.jsonl}. None of the distinctive hook
strings appeared:

\begin{center}
\begin{tabular}{lc}
\toprule
\textbf{Expected hook string} & \textbf{Occurrences} \\
\midrule
\texttt{Face-Card Exposure Metrics} & 0 \\
\texttt{Deck face cards} & 0 \\
\texttt{face-dependent jokers} & 0 \\
\texttt{The Plant boss disables} & 0 \\
\bottomrule
\end{tabular}
\end{center}

Thus, although the source diff was non-empty, the advertised face-exposure
intervention was absent from the model-facing context. The candidate's
development-set gain therefore cannot be attributed to the proposed hook.

This illustrates the noisy-selection failure mode in Balatro. Even with the same
fixed seed and nearly identical prompts, a frontier LLM can produce different
early actions; a single shop purchase, discard, or played hand can change the
future state distribution and final outcome. With only a few long rollouts per
candidate during search, an inert edit can be selected when ordinary rollout
variance makes its score appear better than the reference. This is a
credit-assignment failure: the optimizer observes a higher sparse reward and
attributes it to the candidate diff, while the executed-context audit shows that
the claimed mechanism was inactive.

For long-horizon harness evolution, source-level diffs are therefore
insufficient. Selected edits should be audited at the rendered-context or
execution level to verify that the intended prompt block, hook output, memory
entry, or tool schema is active at the decision states where it is supposed to
apply.
\section{Statistical Validation of Long-Horizon Gains}
\label{app:statistical-analysis}

Under the shared final-evaluation protocol in
Appendix~\ref{app:shared-long-horizon-protocol}, we perform statistical inference on capped
final-round progress in Balatro and the final floor reached in STS2. We treat each seed as a
stratum and its three final-evaluation rollouts as samples of task-execution stochasticity.

For each environment, we compare DemoEvolve with the Base Harness using a two-sided exact
permutation test stratified by seed. Within each seed, we pool the six observations from the
two methods and enumerate all \(\binom{6}{3}=20\) assignments that retain three observations
per method. The exact reference distribution is the Cartesian product of these assignments
across the five seed strata. The test statistic is the aggregate mean difference between
DemoEvolve and the Base Harness over all five seeds. We apply Holm correction across the two
primary environment-level comparisons. We additionally quantify uncertainty in the mean
improvement using seed-stratified percentile bootstrap confidence intervals, resampling the
three rollout observations with replacement within each method and fixed seed.

DemoEvolve improves mean progress over the Base Harness by 5.07 rounds in Balatro
(95\% CI [1.60, 8.67]; exact \(p=.023\); \(p_{\mathrm{Holm}}=.023\)) and by 10.47 floors
in STS2 (95\% CI [6.53, 14.67]; exact \(p=3.94{\times}10^{-5}\);
\(p_{\mathrm{Holm}}=7.88{\times}10^{-5}\)). Both improvements remain significant after
Holm correction.

To more directly isolate the contribution of demonstration guidance, we additionally compare
DemoEvolve against Meta-Harness and Meta-Harness-Web using the same seed-stratified exact
permutation procedure within each environment. For each comparator, we combine the two
environment-level \(p\)-values using Fisher's method and apply Holm correction across the two
resulting cross-environment hypotheses. The combined tests favor DemoEvolve over both
Meta-Harness (\(p_{\mathrm{Holm}}=.013\)) and Meta-Harness-Web
(\(p_{\mathrm{Holm}}=.026\)). These cross-environment tests do not imply significance within
each environment separately.

The DemoEvolve--Base seed-level mean difference is positive for every evaluated seed,
yielding consistent gains in all ten environment--seed blocks. Together, the corrected tests,
confidence intervals excluding zero, and directional consistency support that the selected
DemoEvolve harnesses are robust across the evaluated seeds and task-execution rollout
stochasticity. These analyses condition on the final harness selected by one evolution run per
condition; they do not estimate between-run variance of the outer-loop evolution process.

\FloatBarrier
\section{Construction of In-Context Demonstration Summaries}
\label{sec:appendix-demo-context-summaries}

The raw demonstrations are too long to insert directly into the fixed base harness. We
therefore compress the same nine demonstrations along two complementary axes.
\emph{Episode-level summaries} preserve the internal coherence of an individual
demonstration run, whereas \emph{segment-level summaries} preserve stage-specific resolution by
aggregating demonstrated decisions at corresponding progress states. These summaries define
the two demonstrations-without-evolution baselines reported in the main paper. At inference
time, all nine episode-level summaries are inserted together, while the segment-level baseline
injects only the summary corresponding to the agent's current stage.

\subsection{Summary Generation}

All summaries are generated by \texttt{claude-opus-4-7}, the same model used as the coding
proposer for DemoEvolve, so the summary baselines are not disadvantaged by a weaker
distillation model. Each summary is produced by an independent single-turn call: no state is
carried between calls, and no call observes evaluation rollouts or scores from the
summary-conditioned agent. Each prompt concatenates a summary-specific instruction, a
shared constraint block, and the demonstration material associated with that summary unit.

The episode-level instruction requests a plan of at most 500 words. Its first line identifies
the run's build---which archetype the player committed to and when that commitment
occurred---followed by the chain of decisions supporting that build. The segment-level
instruction also permits at most 500 words but requests only information specific to the
current stage: what should already have been accomplished, which resource levels should be
maintained, and how its tradeoffs differ from adjacent stages. Because demonstrations may
pursue different builds, the instruction asks the model to preserve disagreements and their
preconditions rather than average them into a single unconditional policy.

The shared constraint block prohibits seed-specific answers and requires the exact English
in-game names visible to the agent. For segment-level generation, the prompt also provides a
global summary and asks for only stage-specific information omitted by that global reference,
preventing the segment summaries from collapsing into near-duplicates.

\subsection{Segment Definitions}

Segment boundaries follow each game's native progress structure and are computable directly
from the agent's current state, without retrieval or similarity search. In Balatro, the eight
segments correspond to Antes~1--8. Each ante groups three blinds and their intervening
shops, while score requirements increase across antes; consequently, the expected progress
and resource state are well defined at this granularity.

In STS2, the six segments cover cumulative-floor intervals 1--8, 9--17, 18--25, 26--33,
34--41, and 42+. These intervals split each act near its midpoint, with boundaries aligned to
the Act~1 and Act~2 boss floors at 17 and 33; no boundary divides an ongoing combat. The
harness assigns segments using the same cumulative-floor function used by the benchmark's
progress metric, preventing convention drift between summary injection and evaluation.
\section{Additional Balatro Economy Analysis}
\label{sec:appendix-balatro-economy}

\begin{table}[t]
\centering
\scriptsize
\setlength{\tabcolsep}{2.1pt}
\begin{tabular*}{\columnwidth}{@{\extracolsep{\fill}}lrrrrr@{}}
\toprule
Method & Entry & Late & Reach & Dist. & Late dist. \\
\midrule
Human & 42.99 & 61.87 & 1.00 & 0.00 & 0.00 \\
BalatroLLM & 10.89 & 6.61 & 0.39 & 32.10 & 55.25 \\
Meta-Harness & 12.16 & 6.97 & 0.45 & 30.83 & 54.89 \\
Meta-Harness-Web & 18.96 & 13.51 & 0.32 & 24.03 & 48.36 \\
DemoEvolve & 30.73 & 55.37 & 0.81 & 12.26 & 6.49 \\
\bottomrule
\end{tabular*}
\caption{Attrition-adjusted economy summary. Entry and Late are adjusted entry-money
means overall and over rounds 19--23; Reach is late reach rate. Dist. columns report mean
absolute distance from the human curve; lower is better.}
\label{tab:balatro_human_like_economy}
\end{table}

We compute attrition-adjusted after-blind entry money over all initial rollouts. For method
\(m\), seed panel \(s\), and game round \(r\), let \(N_{m,s}\) denote the number of initial
rollouts. The adjusted mean is
\[
\widetilde{M}_{m,s}(r)
=
\frac{1}{N_{m,s}}
\sum_{i=1}^{N_{m,s}}
\widehat{M}_i(r),
\]
where
\[
\widehat{M}_i(r)
=
\begin{cases}
M_i(r), & \text{if rollout } i \text{ reaches round } r,\\
0, & \text{otherwise}.
\end{cases}
\]
Here, \(M_i(r)\) is the money recorded after entering the blind at round \(r\). Assigning
zero to terminated rollouts prevents late-round economy estimates from being dominated by
the small subset of surviving runs. Thus, late values reflect both survival and economy
management. Curves in the main paper use a three-round moving average for readability;
the numerical summaries in Table~\ref{tab:balatro_human_like_economy} use the unsmoothed
adjusted curves.

\end{document}